\documentclass[lettersize,journal]{IEEEtran}
\usepackage{amsmath,amsfonts}
\usepackage{algorithm}
\usepackage{algpseudocode}
\usepackage{booktabs}
\usepackage{array}
\usepackage[caption=false,font=normalsize,labelfont=sf,textfont=sf]{subfig}
\usepackage{textcomp}
\usepackage{stfloats}
\usepackage{url}
\usepackage{verbatim}
\usepackage{graphicx}
\usepackage{bm}
\usepackage{ragged2e}  

\def\BibTeX{{\rm B\kern-.05em{\sc i\kern-.025em b}\kern-.08em
    T\kern-.1667em\lower.7ex\hbox{E}\kern-.125emX}}
\usepackage{balance}
\begin{document}
\title{StableAnimator++: Overcoming Pose Misalignment and Face Distortion for Human Image Animation}
\author{Shuyuan Tu, Zhen Xing, Xintong Han, Zhi-Qi Cheng, Qi Dai, \IEEEmembership{Member, IEEE}, Chong Luo, \IEEEmembership{Senior Member, IEEE}, \\ Zuxuan Wu, \IEEEmembership{Member, IEEE}, Yu-Gang Jiang, \IEEEmembership{Fellow, IEEE}
\thanks{S. Tu, Z. Xing, Z. Wu, Y-G. Jiang are with School of Computer Science, Fudan University. Email: sytu23@m.fudan.edu.cn, \{zxing20, zxwu\}@fudan.edu.cn}
\thanks{Q. Dai, C. Luo are with Microsoft Research Asia. Emails: \{qid, cluo\}@microsoft.com}
\thanks{X. Han is with Tencent Inc. Emails: pathan@tencent.com}
\thanks{Z. Cheng is with University of Washington. Emails: zhiqics@uw.edu}
}

\markboth{Journal of \LaTeX\ Class Files,~Vol.~18, No.~9, September~2020}%
{How to Use the IEEEtran \LaTeX \ Templates}

\maketitle

\begin{abstract}

Current diffusion models for human image animation often struggle to maintain identity (ID) consistency, especially when the reference image and driving video differ significantly in body size or position.
We introduce StableAnimator++, the first ID-preserving video diffusion framework with learnable pose alignment, capable of generating high-quality videos conditioned on a reference image and a pose sequence without any post-processing.
Building upon a video diffusion model, StableAnimator++ contains carefully designed modules for both training and inference, striving for identity consistency. 
In particular, StableAnimator++ first uses learnable layers to predict the similarity transformation matrices between the reference image and the driven poses via injecting guidance from Singular Value Decomposition (SVD). 
These matrices align the driven poses with the reference image, mitigating misalignment to a great extent.
StableAnimator++ then computes image and face embeddings using off-the-shelf encoders, refining the face embeddings via a global content-aware Face Encoder. To further maintain ID, we introduce a distribution-aware ID Adapter that counteracts interference caused by temporal layers while preserving ID via distribution alignment.
During the inference stage, we propose a novel Hamilton-Jacobi-Bellman (HJB) based face optimization integrated into the denoising process, guiding the diffusion trajectory for enhanced facial fidelity.
Experiments on benchmarks show the effectiveness of StableAnimator++ both qualitatively and quantitatively.
Project website: \url{https://francis-rings.github.io/StableAnimator++/}.
\end{abstract}


\section{Introduction}
\IEEEPARstart{H}{uman} image animation~\cite{wang2024disco, xu2024magicanimate, hu2024animate, zhu2024champ, wang2024unianimate, zhang2024mimicmotion, peng2024controlnext, tan2024animate_x, tu2022multiple, tu2023implicit} aims to animate a reference image based on the motion pattern of a pose sequence, enabling diverse applications in entertainment and virtual reality. The phenomenal successes of diffusion models~\cite{dhariwal2021diffusion,ho2020denoising,ho2022cascaded,song2020score,song2020denoising,rombach2022high,meng2021sdedit,tumanyan2023plug,weng2024genrec,xing2024survey,xing2024simda,xing2024aid,magicmotion} in video generation significantly inspire the advancement of human image animation.
However, when dealing with pose sequences that exhibit significant motion variation, current approaches suffer from substantial distortions and inconsistencies, particularly in facial regions, destroying ID information.
Misalignment in body size and position between the reference image and the driving video, which is common in real-world applications, further exacerbates this issue.

To address this issue, there are numerous methods exploring identity (ID) preservation~\cite{ye2023ip-adapter, wang2024instantid, huang2024consistentid, guo2024pulid} for image generation, yet limited effort has been made for videos.
While one could add temporal modeling layers to image diffusion models, doing so would inevitably disrupt the original spatial priors essential for identity preservation. Since image-based ID-preserving methods depend on these stable priors, introducing temporal layers often leads to poor results. This makes maintaining identity while ensuring video quality a major challenge for image animation.
Furthermore, recent animation models~\cite{zhang2024mimicmotion, peng2024controlnext} rely on FaceFusion~\cite{facefusion} for post-processing, which also degrades the quality of animated videos, particularly for facial areas.

\begin{figure}[t!]
\begin{center}
\includegraphics[width=1\linewidth]{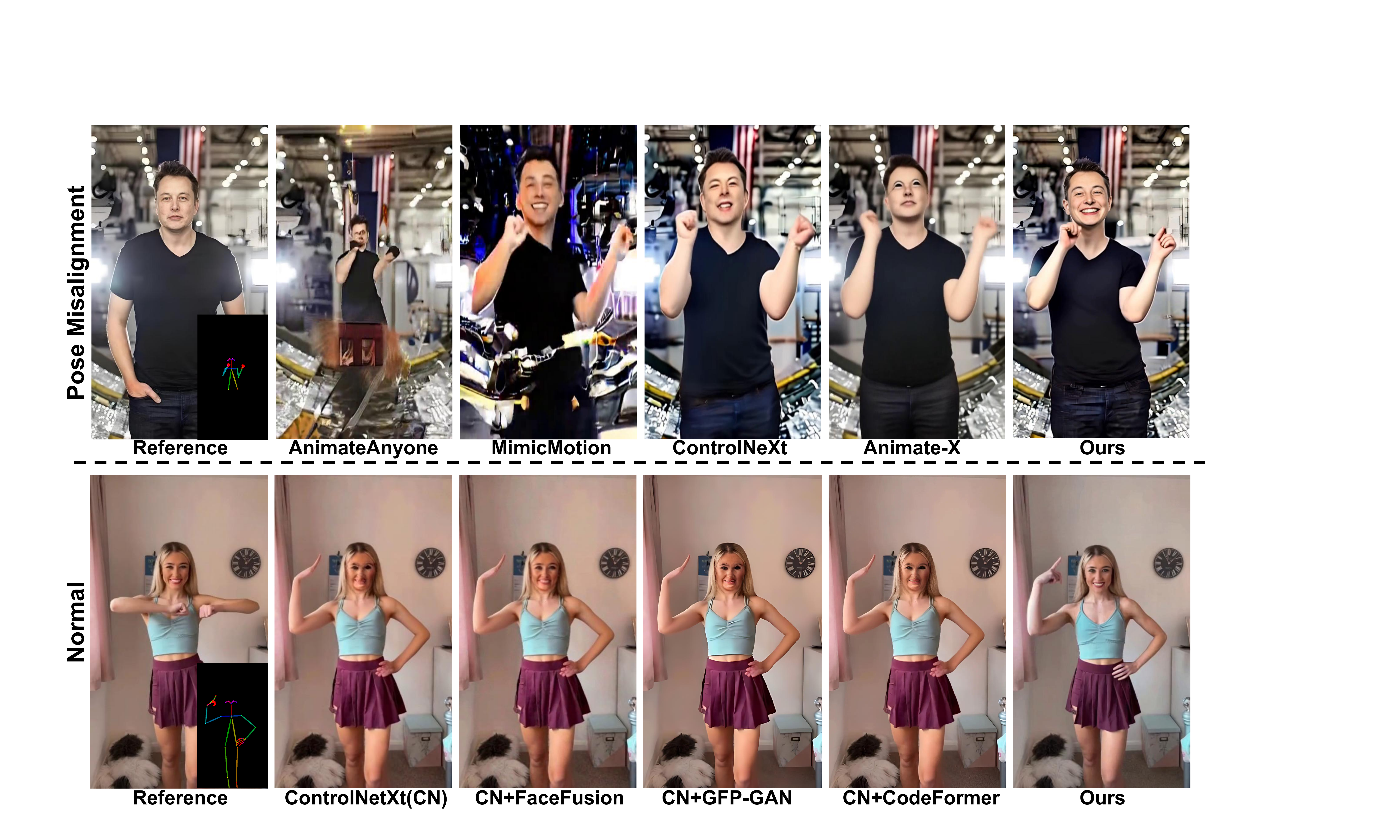}
\end{center}
\vspace{-0.3cm}
   \caption{Pose-driven Human image animations generated by our StableAnimator++ and compared methods, showing its power to synthesize ID-preserving videos even in scenarios with significant pose misalignment between the reference and driven poses. AnimateAnyone~\protect\cite{hu2024animate}, MimicMotion~\protect\cite{zhang2024mimicmotion}, ControlNeXt~\protect\cite{peng2024controlnext}, and Animate-X~\protect\cite{tan2024animate_x} are existing open-source animation models.
   FaceFusion~\protect\cite{facefusion} is a face-swapping tool. GFP-GAN~\protect\cite{wang2021gfpgan} and CodeFormer~\protect\cite{zhou2022codeformer} are face restoration models.
   Normal refers to the pose-aligned scenario.
   }
\label{fig:cover}
\vspace{-0.50cm}
\end{figure}

Regarding the pose misalignment, previous methods~\cite{tu2024motioneditor, wang2024unianimate, zhang2024mimicmotion, peng2024controlnext} utilize a pose alignment algorithm to align the driven pose with the reference image before animation, which roughly calculates the scaling factor and offset based on the relative size ratio between the reference image and the driven pose to scale and translate the driven pose. Champ~\cite{zhu2024champ} leverages the parametric shape alignment to align the 3D signal SMPL.
However, in scenarios with significant discrepancies in body size and protagonist's position, these approaches become highly inaccurate, negatively impacting the quality of the animated video.
Furthermore, while Animate-X~\cite{tan2024animate_x} claims to be insensitive to body size and protagonist's position gaps between the reference image and driven poses, our experiments show that dramatic pose misalignment still significantly degrades the quality of animations in such scenarios.

In light of this, we propose StableAnimator++, consisting of dedicated modules for both training and inference to maintain ID consistency for high-quality human image animation in various scenarios, including dramatic pose misalignment.
StableAnimator++ first introduces learnable layers to predict similarity transformation matrices (rotation, scaling, and translation) between the reference image and driven poses, guided by Singular Value Decomposition (SVD). Since directly predicting aligned poses is challenging, SVD provides an intermediate transformation state to guide the learnable layers via cross-attention, significantly enhancing the model's ability to capture the projection relationship between the reference image and driven poses. Trained layers offer greater robustness and accuracy in alignment in various scenarios than conventional methods. 
It uses the similarity transformation matrices to align driven poses with the reference image, reducing gaps in body size and protagonist position.
Then, StableAnimator++ uses off-the-shelf extractors~\cite{deng2019arcface, radford2021learning} to obtain face and image embeddings for the reference image, respectively. 
Face embeddings are further refined by a global content-aware Face Encoder to enable interaction with the reference, enhancing face embeddings' perception of the reference's overall layout, such as backgrounds.
The refined face embeddings are fed to a video diffusion model with a novel distribution-aware ID Adapter that ensures video fidelity while preserving ID clues. 
In particular, diffusion latents perform separate cross-attention with refined face and image embeddings, respectively, with their means and variances computed. 
We then use respective means and variances to conduct the distribution alignment between the resulting outputs. It effectively mitigates interference from the temporal layers by progressively bringing two distributions closer at each step, ensuring ID consistency without compromising video fidelity.

During inference, to further enhance face quality and reduce reliance on post-processing tools, StableAnimator solves
the Hamilton-Jacobi-Bellman (HJB) equation~\cite{bardi1997optimal, peng1992stochastic} for face optimization. 
We find that solving the HJB equation corresponds with the core principles of diffusion denoising. Therefore, we incorporate the HJB equation into the inference process, which allows a controllable variable to guide and constrain the direction of the denoising process. In particular, the solution of HJB is used to update the latents for each denoising step, constraining the denoising path and directing the model toward optimal ID consistency. Since this procedure always adapts to the current distribution of denoised latents, the simultaneous denoising and face optimization effectively eliminates detail distortions. Thus, it can replace the previous over-reliance on third-party post-processing tools, such as face-swapping tools.

As shown in Fig. \ref{fig:cover}, while Animate-X~\cite{tan2024animate_x} suffers from dramatic body distortion, StableAnimator++ can effectively animate the reference image based on the pose sequence in the significant pose misalignment scenario. In the normal scenario, while ControlNeXt~\cite{peng2024controlnext} exhibits severe facial and body distortions despite using face swapping or restoration tools, StableAnimator++ can accurately animate the reference based on given poses while preserving ID consistency.

In conclusion, our contributions are as follows:
(1) We propose a novel learnable SVD-guided pose alignment model, which takes scaling, rotation, and translation into account, significantly reducing gaps from misalignment issues. To our knowledge, we are the first to explore learnable pose alignment for ID-preserving human image animation across various scenarios.
(2) We propose a global content-aware Face Encoder and a novel distribution-aware ID Adapter to enable the video diffusion model to incorporate face embeddings without compromising video fidelity.
(3) We propose a novel HJB equation-based face optimization method that further enhances face quality while conducting conventional denoising. It is only active in the inference without training any diffusion components.
(4) Experimental results on benchmark datasets show the superiority of our model over the SOTA.

A preliminary version of this paper appeared in~\cite{tu2024stableanimator}. The present paper includes a complete literature review on robust human image animation models, with a focus on handling pose misalignment commonly observed in real-world applications; an updated solution that utilizes learnable layers to predict similarity transformation matrices (rotation, scaling, and translation) between the reference image and driven poses, guided by Singular Value Decomposition.

\section{Related Work}
\noindent \textbf{Diffusion for Video Generation.}
Diffusion models have achieved remarkable success in video generation~\cite{dhariwal2021diffusion,meng2021sdedit,nichol2021improved,hertz2022prompt,ho2020denoising,ho2022cascaded,song2020denoising,song2020score,tumanyan2023plug}, driven by their superior diversity and high fidelity. Current video generation models~\cite{singer2022make, guo2023animatediff, wu2023tune, wang2024magicvideo, tu2024motioneditor, videoworldsimulators2024,tu2024motionfollower,tu2024stableanimator} capture spatio-temporal representations by adding temporal layers to pre-trained image generation models. Some works~\cite{peebles2023scalable, yan2021videogpt, yu2023magvit, ma2024latte, bao2024vidu, hong2022cogvideo, kong2024hunyuanvideo} replace the U-Net with transformers to scale up, showing a significant advancement in large video generation models. Following recent animation models~\cite{peng2024controlnext, zhang2024mimicmotion}, we adopt Stable Video Diffusion~\cite{blattmann2023stable} as the backbone.

\vspace{0.2cm}
\noindent \textbf{Pose-guided Human Image Animation.}
Human image animation transfers motion from a given pose sequence to a reference image. Early works~\cite{Siarohin_2019_NeurIPS, siarohin2021motion, huang2021few} primarily relied on GANs~\cite{goodfellow2020generative}, but GAN-based models often suffer from flickering issues. Sparked by the diffusion models in video generation, recent animation models are basically based on diffusion models. 
Disco~\cite{wang2024disco} is the first to try the diffusion model in human animation. MagicAnimate~\cite{xu2024magicanimate} and AnimateAnyone~\cite{hu2024animate} both introduce transformer-based temporal attention modules for temporal smoothness. Champ~\cite{zhu2024champ} uses 3D signal SMPL to model motion patterns. Unianimate~\cite{wang2024unianimate} inserts Mamba~\cite{mamba2} into the diffusion U-Net for efficiency. MimicMotion~\cite{zhang2024mimicmotion} introduces the regional loss to address hand distortion. ControlNext~\cite{peng2024controlnext} proposes a convolution-based PoseNet. Animate-X~\cite{tan2024animate_x} aims to animate various character types. However, previous animation models suffer from face distortion. As they utilize the third-party face-swapping tool FaceFusion~\cite{facefusion} as post-processing to address this issue, yet this approach can degrade overall video quality. This issue becomes more severe when there is a misalignment in body size and position between the reference image and the driven pose. Our StableAnimator++ can still synthesize ID-preserving videos even when confronting dramatic pose misalignment scenarios without relying on any post-processing tools.

\vspace{0.2cm}
\noindent\textbf{ID Consistency Image Generation.}
Recent studies have explored identity (ID) preservation in the image domain. LoRA~\cite{hu2021lora} injects a few trainable parameters for personalized tuning but requires separate training for each identity, limiting scalability. IP-Adapter-FaceID~\cite{ye2023ip-adapter} decouples cross-attention for text and facial features, potentially causing feature misalignment. PhotoMaker~\cite{li2024photomaker}, FaceStudio~\cite{yan2023facestudio}, and InstantID~\cite{wang2024instantid} refine facial embeddings through hybrid mechanisms, while ConsistentID~\cite{huang2024consistentid} leverages a facial prompt generator for detail preservation. PuLID~\cite{guo2024pulid} introduces contrastive and ID-specific losses to enhance identity fidelity. However, these approaches are not readily compatible with video diffusion models, where temporal layers may disrupt spatial consistency, leading to domain mismatch and degraded animation quality. In contrast, our StableAnimator++ integrates ID information into video diffusion models via a distribution-aware ID Adapter, effectively resolving the conflict between ID consistency and video fidelity.

\begin{figure*}[t!]
\begin{center}
\includegraphics[width=1\linewidth]{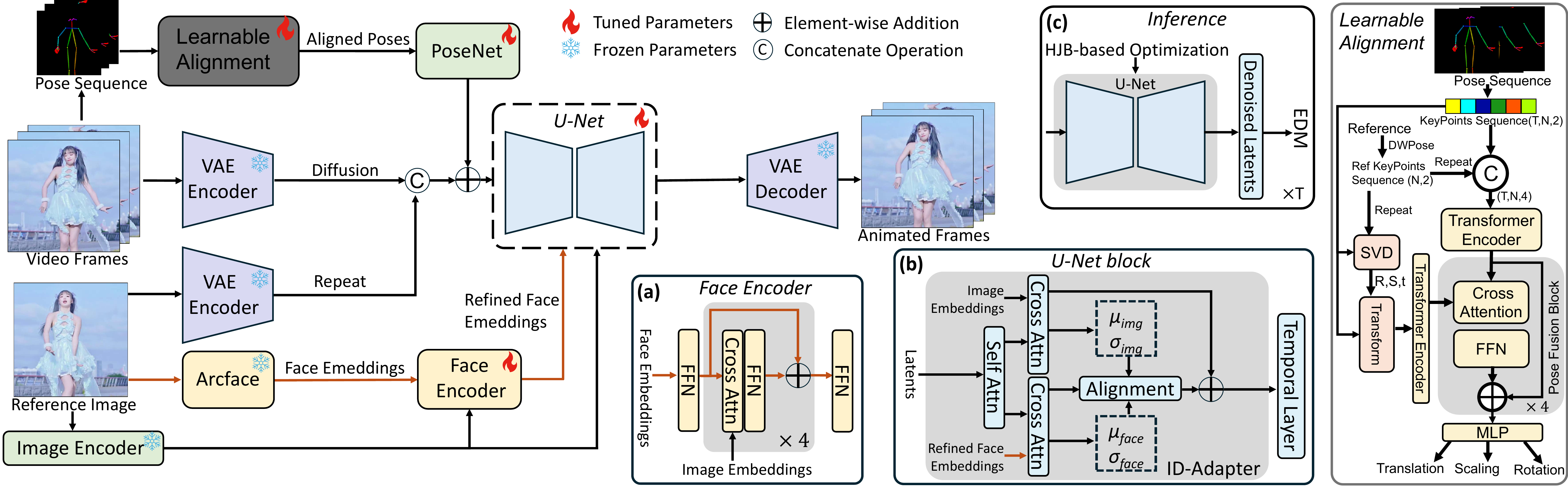}
\end{center}
\vspace{-0.6cm}
   \caption{Architecture of StableAnimator++. (a) and (b) refer to the structure of the Face Encoder and each block in the U-Net. 
   We first apply our learnable alignment to the driving pose sequence and feed the aligned results into the PoseNet for motion modeling.
   Embeddings from the Image Encoder and Face Encoder are injected into each block of U-Net. Given the reference, we extract the image embeddings and face embeddings utilizing Image Encoder and Arcface. 
   The face embeddings are fed into the FaceEncoder to enhance ID. Then, image embeddings and refined face embeddings are injected into the U-Net through the ID Adapter to ensure ID consistency.
   }
\label{fig:framework}
\vspace{-0.5cm}
\end{figure*}

\section{Method}

As shown in Fig. \ref{fig:framework}, inspired by previous works~\cite{zhang2024mimicmotion, peng2024controlnext}, StableAnimator++ is based on the commonly used Stable Video Diffusion~\cite{blattmann2023stable}. 
The driven pose sequence is aligned using our learnable alignment block and then processed by a PoseNet, as depicted in Sec. \ref{sec: pose_alignment}. A PoseNet with a similar architecture to AnimateAnyone~\cite{hu2024animate} encodes the aligned poses, which are then added to the noisy latents.
A reference image is fed to the diffusion model through three pathways: (1) Converted into a latent code using a frozen VAE Encoder~\cite{kingma2013auto}. This latent code is then duplicated to align with the number of video frames and concatenated with the diffusion latents. (2) Encoded by the CLIP Image Encoder~\cite{radford2021learning} to obtain image embeddings, which are then fed to each cross-attention block of a denoising U-Net, guiding the synthesized appearance. (3) Input to Arcface~\cite{deng2019arcface} to gain face embeddings, which are subsequently refined for further alignment via our Face Encoder. Refined face embeddings are then fed to the denoising U-Net. More details are described in Sec. \ref{sec: face_consistency}.

We replace the original input video frames with random noise during inference, while the other inputs stay the same. We propose a novel HJB-equation-based face optimization to enhance ID consistency and eliminate reliance on third-party post-processing tools.
It integrates the solution process of the HJB equation into the denoising, allowing optimal gradient direction toward high ID consistency as detailed in Sec. \ref{sec: optimization}.

\subsection{Learnable Alignment During Training}
\label{sec: pose_alignment}
Previous pose alignment methods~\cite{tu2024motioneditor, wang2024unianimate, zhang2024mimicmotion, peng2024controlnext} in animation basically calculate the scaling factor and offset~\cite{tu2024motioneditor} based on the relative size ratio between the reference image and the driven pose to adjust the driven skeleton keypoints. Champ~\cite{zhu2024champ} utilizes the parametric shape alignment to align the 3D signal SMPL. The above approaches are particularly inaccurate in cases of significant body size misalignment or positional discrepancies between the reference image and the driven video, thereby degrading the animation quality. While Animate-X~\cite{tan2024animate_x} claims to be pose-agnostic and alignment-free, it still suffers from body distortions in cases of significant misalignment. To address this, we introduce a novel learnable alignment that uses learnable layers to predict accurate similarity transformation matrices (rotation, scaling, and translation) between the reference image and driven videos, guided by Singular Value Decomposition (SVD).
Employing learnable layers to predict the aligned poses is relatively more effective and robust compared with conventional methods, as it is trained on diverse misalignment scenarios.

Fig. \ref{fig:framework} illustrates the overall framework of our alignment block.
Given a reference image $\bm{Ref}\in{\cal{R}}^{3 \times H \times W}$ and a driven video $\bm{V}\in{\cal{R}}^{T \times 3 \times H \times W}$, we leverage DWPose~\cite{yang2023effective} to extract their pose keypoint sequences $\bm{P}_{r}\in{\cal{R}}^{2 \times N}$ and $\bm{P}_{d}\in{\cal{R}}^{2 \times N \times T}$, respectively. $N$ refers to the keypoint number, with a default value of 18 in DWPose. $T$ is the frame number of the driven video. We first repeat $\bm{P}_{r}$ to obtain $\bm{P}_{r}^{*}$ and concatenate them with $\bm{P}_{d}$ along the channel dimension, then feed them into a Transformer Encoder $\mathtt{Encoder}_{m}(\cdot)$ to model their motion patterns and relative positional relationships:
\begin{equation}\small
\label{eq:motion_modeling}
\begin{aligned}
     \bm{F}_{m}=\mathtt{Encoder}_{m}(\mathtt{Concat}(\mathtt{Repeat}(\bm{P}_{r}), \bm{P}_{d})),
\end{aligned}
\end{equation}
where $\mathtt{Concat}(\cdot)$ is the concatenation operation. 
Furthermore, inspired by ICP~\cite{besl1992method}, we use SVD to obtain intermediate aligned poses, guiding the learnable layers to model the projection relationship between $\bm{P}_{r}$ and $\bm{P}_{d}$, as directly predicting the aligned keypoints is challenging for the learnable layers. 
ICP, designed for point clouds, iteratively optimizes transformation matrices without considering point correspondences (\emph{e.g.}, hand to hand), and its accuracy is unstable, making it impractical for animation.
Thus, we use SVD only once in alignment as guidance for an intermediate state. In particular, we center $\bm{P}_{r}^{*}$ and $\bm{P}_{d}$ as follows:
\begin{equation}\small
\label{eq:center}
\begin{aligned}
     \bm{X}_{r}=\bm{P}_{r}^{*}-\frac{1}{N}\sum_{i=1}^{N}\bm{P}_{r}^{*}[:, i, :], 
     \bm{X}_{d}=\bm{P}_{d}-\frac{1}{N}\sum_{i=1}^{N}\bm{P}_{d}[:, i, :].
\end{aligned}
\end{equation}
Notably, we describe the case where only one person is present in the input image for readability.
To determine the optimal rotation matrix $\bm{R}$, we first construct the covariance matrix $\bm{K}$, which captures the correlation between the two centered point sets:
\begin{equation}\small
\label{eq:covariance}
\begin{aligned}
     \bm{K}=\bm{X}_{d}\bm{X}_{r}^{T}.
\end{aligned}
\end{equation}
We then apply Singular Value Decomposition ($\mathtt{SVD}(\cdot)$) to decouple $\bm{K}$ as follows:
\begin{equation}\small
\label{eq:svd}
\begin{aligned}
     \bm{U}, \bm{s}, \bm{V}^{T}=\mathtt{SVD}(\bm{K}),
\end{aligned}
\end{equation}
where orthogonal matrices $\bm{U}$ and $\bm{V}^{T}$ describe the principal axes of variation. We can obtain the rotation matrix $\bm{R}$:
\begin{equation}\small
\label{eq:rotation}
\begin{aligned}
     \bm{R}=\bm{V}\bm{U}^{T}.
\end{aligned}
\end{equation}
Furthermore, we use $\bm{R}$ to obtain the scale factor:
\begin{equation}\small
\label{eq:scaling}
\begin{aligned}
     \bm{S}=\frac{\mathtt{Trac}(\bm{R}\bm{K})}{\sum_{i=1}^{N}{\bm{X}_{d}^{i}}^2},
\end{aligned}
\end{equation}
where $\mathtt{Trac}(\cdot)$ and $\sum_{i=1}^{N}{\bm{X}_{d}^{i}}^2$ refer to the trace operator and the dispersion of the body shape of the driven frame in space. The translation vector $\bm{t}$ describes the displacement between the centroids of the reference body and the driven frame body after rotation and scaling as:
\begin{equation}\small
\label{eq:translation}
\begin{aligned}
     \bm{t}=\frac{1}{N}\sum_{i=1}^{N}\bm{P}_{r}^{*}[:, i, :]-\bm{S}*(\bm{R}\frac{1}{N}\sum_{i=1}^{N}\bm{P}_{d}[:, i, :]).
\end{aligned}
\end{equation}
We can further use the above $\bm{R}$, $\bm{S}$, and $\bm{t}$ to transform the initial driven keypoints sequence $\bm{P}_{d}$ into an intermediate state $\bar{\bm{P}_{d}}$ as follows:
\begin{equation}\small
\label{eq:intermediate_state}
\begin{aligned}
     \bar{\bm{P}_{d}}=\bm{S}*(\bm{R}\cdot\bm{P}_{d})+\bm{t}.
\end{aligned}
\end{equation}
We then input $\bar{\bm{P}_{d}}$ to another Transformer Encoder $\mathtt{Encoder}_{svd}(\cdot)$ to extract motion-aware features, which are subsequently passed to a Pose Fusion Block $\mathtt{PFusion}(\cdot)$ for guidance injection as follows:
\begin{equation}\small
\label{eq:guidance}
\begin{aligned}
     \bar{\bm{F}_{m}}=\mathtt{PFusion}(\mathtt{Encoder}_{svd}(\bar{\bm{P}_{d}}), \bm{F}_{m}),
\end{aligned}
\end{equation}
where $\mathtt{PFusion}(\cdot)$ contains 4 modules, each comprising a cross-attention layer and an FFN. Although the SVD output from a single interaction may not be strictly accurate, injecting this guidance into the main features via cross-attention still significantly enhances the model's ability to capture discrepancies in body size and position between the reference and driven poses, thereby facilitating their learning. We then use an MLP to predict the rotation/scaling/translation matrices ($\bm{R}^{'},\bm{S}^{'},\bm{t}^{'}$) as follows:
\begin{equation}\small
\label{eq:guidance}
\begin{aligned}
     \bm{R}^{'},\bm{S}^{'},\bm{t}^{'}=\mathtt{MLP}(\bar{\bm{F}_{m}}).
\end{aligned}
\end{equation}
The above operation is set as $\bm{R}^{'}_{a},\bm{S}^{'}_{a},\bm{t}^{'}_{a}$=$\mathtt{Align}(\bm{P}_{a}, \bm{P}_{b})$, where $\bm{P}_{a}$ is the keypoints to be aligned and $\bm{P}_{b}$ is the reference keypoints. The ultimate aligned driven poses $\bm{P}_{d}^{align}$ can be obtained as follows via applying $\mathtt{Align}(\bm{P}_{d}, \bm{P}_{r})$:
\begin{equation}\small
\label{eq:align}
\begin{aligned}
     \bm{P}_{d}^{align}=\bm{S}^{'}_{d}*(\bm{R}^{'}_{d}\cdot\bm{P}_{d})+\bm{t}^{'}_{d}.
\end{aligned}
\end{equation}

We train the alignment block from scratch at the image level for 50 epochs using 5K collected videos before training the entire StableAnimator++.
With an average video length of 60 seconds and 30 FPS, the total number of training images exceeds 9 M.
We first select two frames from a training video: one as the reference image and the other as the driven pose.
For each driven pose, we modify it by applying random scaling, rotation, and translation matrices to simulate misalignment. We then feed the modified driven pose $\bm{P}_{d}$ and the reference image to our alignment block for predicting accurate transformation matrices ($\bm{R}^{'}_{d},\bm{S}^{'}_{d},\bm{t}^{'}_{d}$). We calculate the average Euclidean distance $\mathtt{Dis}_{Euc}(\cdot)$ between aligned poses and ground-truths $\bm{P}_{d}^{gt}$ as the loss function:
\begin{equation}\small
\label{eq:align_loss}
\begin{aligned}
     \bm{L}_{align}=\mathtt{Avg}(\mathtt{Dis}_{Euc}(\bm{P}_{d}^{gt}, \bm{S}^{'}_{d}*(\bm{R}^{'}_{d}\cdot\bm{P}_{d})+\bm{t}^{'}_{d})).
\end{aligned}
\end{equation}

\subsection{ID-preserving During Training}
\label{sec: face_consistency}

\noindent\textbf{Global Content-aware Face Encoder.}
To synthesize ID-preserving animations guided by a pose sequence, it's essential to retain both the facial details and the global context of the reference image. 
Although directly injecting face embeddings into the U-Net enhances facial fidelity, it fails to capture the global context (layout and background) in the reference image before being injected into the U-Net. Consequently, ID-irrelevant elements in the reference image bring noise to face modeling, impairing the overall animation quality.
To overcome this, we introduce a Global Content-Aware Face Encoder, which refines face embeddings by allowing them to interact with the full reference image through a series of cross-attention blocks, enabling more context-aware modeling as shown in Fig. \ref{fig:framework}.

\vspace{0.1cm}
\noindent\textbf{Distribution-aware ID Adapter.}
To mitigate the distortion of spatial features occurring when directly incorporating image-domain ID-preserving methods~\cite{Siarohin_2019_NeurIPS, guo2024pulid, huang2024consistentid, wang2024instantid} into the video diffusion model, the outputs of the Face Encoder are further fed to our ID Adapter.
Feature distortion describes the misalignment between face embeddings and spatial diffusion latents, caused by distribution shifts when temporal layers are added at each denoising step.
Image-domain ID-preserving methods rely heavily on a stable spatial distribution of diffusion latents, but temporal layers often alter this distribution, leading to instability in ID preservation.
This results in a conflict between preserving high video fidelity and maintaining identity integrity, often manifesting as facial blurring or background degradation in the animations. 
As shown in Fig. \ref{fig:framework} (b), our Distribution-aware ID Adapter is incorporated into each spatial layer of the U-Net. It performs distribution alignment between refined face embeddings and diffusion latents before each temporal modeling, effectively mitigating feature distortion.

Concretely, following the standard operation of spatial layers in the diffusion model, we first apply spatial self-attention on latents $\bm{z}_{i}$. The latents of the U-Net perform cross-attention with image embeddings $\bm{emb}_{img}$ and refined face embeddings $\bm{emb}_{face}$, respectively:
\begin{equation}\small
\label{eq:cross_attention}
\begin{aligned}
     \bm{z}_{i}&=\mathtt{SAttn}(\bm{z}_{i}), \\
     \bm{z}^{img}_{i}&=\mathtt{CAttn}(\bm{z}_{i}, \bm{emb}_{img}), \\
     \bm{z}^{face}_{i}&=\mathtt{CAttn}(\bm{z}_{i}, \bm{emb}_{face}),
\end{aligned}
\end{equation}
where $\mathtt{SAttn}(\cdot)$ and $\mathtt{CAttn}(\cdot)$ refer to self-attention and cross-attention operations. 
To align $\bm{z}^{img}_{i}$ and $\bm{z}^{face}_{i}$, we enforce $\frac{\bm{z}^{img}_{i}-\bm{\mu}_{img}}{\bm{\sigma}_{img}}=\frac{\bm{z}^{face}_{i}-\bm{\mu}_{face}}{\bm{\sigma}_{face}}$, where $\bm{\mu}_{img/face}$ and $\bm{\sigma}_{img/face}$ refer to the mean and standard deviation of $\bm{z}^{img/face}_{i}$, respectively. If the equation above holds, the feature distributions on both sides are basically in the same domain. Thus, the aligned $\bm{z}^{face}_{i}$ is element-wise added to $\bm{z}^{img}_{i}$ for maintaining ID consistency:
\begin{equation}\small
\label{eq:alignment}
\begin{aligned}
     \bar{\bm{z}}^{face}_{i}&=\frac{\bm{z}^{face}_{i}-\bm{\mu}_{face}}{\bm{\sigma}_{face}}\times\bm{\sigma}_{img}+\bm{\mu}_{img}, \\
     \bar{\bm{z}_{i}}&=\bar{\bm{z}}^{face}_{i}+\bm{z}^{img}_{i}.
\end{aligned}
\end{equation}
The outputs of our ID Adapter $\bar{\bm{z}_{i}}$ are then fed to temporal layers for temporal modeling. 
When spatial distribution is altered by temporal layers, the aligned $\bar{\bm{z}}^{face}_{i}$ remains in the same domain as $\bm{z}^{img}_{i}$, enabling the original $\bm{z}^{face}_{i}$ to reduce reliance on the unstable spatial distribution. Thus, temporal modeling does not impede the ID information in the U-Net.

\begin{algorithm}[t!]
\caption{Face Optimization ($\sigma(t) = t$ and $s(t) = 1$)}
\label{alg:face_optimization}
\begin{algorithmic}
\small 
\State \textbf{Input:} {$\mathtt{D}_{\theta}(\bm{x}; \bm{\sigma}),t_{i\in \{0, \ldots, N\}}, \bm{\gamma}_{i\in \{0, \ldots, N-1\}}, \bm{y}$} 
    \State \textbf{Sample} $\bm{x}_0 \sim \mathcal{N}(0, t_0^2\bm{I})$ \hfill $\triangleright$ $\mathtt{D}_{\theta}(\bm{x}; \bm{\sigma})$ is a diffusion model
    \State \textbf{For} $i \in \{0, \ldots, N-1\}$ \textbf{do} \hfill $\triangleright$ $t_{i\in \{0, \ldots, N\}}$ are timesteps
        \State \hspace{1em} $\bm{\gamma}_i = 0$ \hfill $\triangleright$ $\bm{\gamma}_{i\in \{0, \ldots, N-1\}}$ are pre-defined factors.
        \State \hspace{1em} \textbf{if} $t_i \in [\bm{S}_{t_{\text{min}}}, \bm{S}_{t_{\text{max}}}]:$ \hfill $\triangleright$ $\bm{y}$ is the reference image.
        \State \hspace{1em} \hspace{1em} $\bm{\gamma}_i = \min \left( \frac{\bm{S}_{\text{churn}}}{N}, \sqrt{2}-1 \right)$
        \State \hspace{1em} \textbf{Sample} $\bm{\epsilon}_i \sim \mathcal{N}(0, \bm{S}_{\text{noise}}^2\bm{I})$
        \State \hspace{1em} $\hat{t}_i = t_i + \bm{\gamma}_i t_i$
        \State \hspace{1em} $\hat{\bm{x}}_i = \bm{x}_i + \sqrt{\hat{t}_i^2 - t_i^2} \bm{\epsilon}_i$
        \State \hspace{1em} $\bm{x}_{\text{pred}}=\mathtt{D}_{\theta}(\hat{\bm{x}}_i; \hat{t}_i)$
        \State \hspace{1em} $\bm{x}_{\text{op}}=\bm{x}_{\text{pred}}.\mathtt{clone}().\mathtt{detach}()$ \hfill $\triangleright$ Starting optimization 
        \State \hspace{1em} $\bm{op}=\mathtt{Adam}([\bm{x}_{\text{op}}], \bm{\eta})$ \hfill $\triangleright$ $\mathtt{Adam}$ optimizer
        \State \hspace{1em} $\bm{x}_{\text{op}}.\text{requires\_grad}=\text{True}$ \hfill $\triangleright$ $\bm{x}_{\text{op}}$ is a HJB variable
        \State \hspace{1em} \textbf{For} $k \in \{1,2, \ldots, 10\}$ \textbf{do} \hfill $\triangleright$ $k$ is the optimization step
        \State \hspace{1em} \hspace{1em} $\bm{f}_{\text{pred}}=\mathtt{Decoder}(\bm{x}_{\text{op}})$ \hfill $\triangleright$ $\mathtt{Decoder}$ is a VAE decoder
        \State \hspace{1em} \hspace{1em} $\bm{loss}=(1-\mathtt{Cos}(\mathtt{Arc}(\bm{f}_{\text{pred}}), \mathtt{Arc}(\bm{y}))).\text{abs}().\text{mean}()$
        \State \hspace{1em} \hspace{1em}  $\bm{op}.\text{zero\_grad}()$
        \State \hspace{1em} \hspace{1em}  $\bm{loss}.\text{backward}(\text{retain\_graph=True})$
        \State \hspace{1em} \hspace{1em}  $\bm{op}.\text{step}()$
        \State \hspace{1em} $\bm{x}_{\text{pred}}=\bm{x}_{\text{op}}$ \hfill $\triangleright$ End of Optimization 
        \State \hspace{1em} $\bm{d}_i = (\hat{\bm{x}}_i-\bm{x}_{\text{pred}})/\hat{t}_i$
        \State \hspace{1em} $\bm{x}_{i+1} = \hat{\bm{x}}_i + (t_{i+1} - \hat{t}_i)\bm{d}_i$
        \State \hspace{1em} \textbf{if} $t_{i+1} \neq 0$:
        \State \hspace{1em} \hspace{1em} $\bm{d}'_i = (\bm{x}_{i+1}-\mathtt{D}_{\theta}(\bm{x}_{i+1}; t_{i+1}))/t_{i+1}$
        \State \hspace{1em} \hspace{1em} $\bm{x}_{i+1} = \hat{\bm{x}}_i + (t_{i+1} - \hat{t}_i) \left( \frac{1}{2} \bm{d}_i + \frac{1}{2} \bm{d}'_i \right)$
    \State \textbf{return} $\bm{x}_N$
\end{algorithmic}
\end{algorithm}

\subsection{ID-preserving During Inference}
\label{sec: optimization}
To improve ID consistency, recent animation works~\cite{zhang2024mimicmotion, peng2024controlnext} use a third-party face-swapping tool FaceFusion~\cite{facefusion}, for post-processing faces. However, animations suffer from overall quality degradation due to excessive reliance on post-processing tools. The reason is that post-processing tools can disrupt the original pixel distribution, as faces generated by third-party tools are not aligned with the domain of the original animations. 
To address this issue, inspired by the HJB equation~\cite{bardi1997optimal, peng1992stochastic, chen2023generative}, we propose the HJB Equation-based Face Optimization.
The HJB equation guides optimal variable selection at each moment in a dynamic system to maximize the cumulative reward.
In our setting, this reward refers to ID consistency, which we aim to enhance by integrating the HJB equation with the diffusion denoising process.
The variable refers to the predicted sample by the diffusion model at each denoising iteration.
We first introduce the process of our face optimization and then demonstrate its rationale.

In particular, we optimize the predicted sample $\bm{x}_{\text{pred}}$ by minimizing the face similarity distance between $\bm{x}_{\text{pred}}$ and the reference before employing denoising (EDM~\cite{karras2022elucidating}) at each step. 
The details are in the Algorithm \ref{alg:face_optimization}, following the structure of the Algorithm 2 in the EDM paper~\cite{karras2022elucidating}. 
$\bm{S}_{\text{noise}}$, $ \bm{S}_{\text{churn}}$,  $\bm{S}_{t_{\text{min}}}$ and 
$\bm{S}_{t_{\text{max}}}$ are the pre-defined values of EDM. $\mathtt{Arc}(\cdot)$ and $\bm{\eta}$ are Arcface~\cite{deng2019arcface} and a learning rate. We employ our optimization to refine the prediction of the diffusion regarding the face similarity with the reference.

The optimized $\bm{x}_{\text{pred}}$ can steer the denoising process forward in a way that maximizes ID consistency. As our optimization relies on the current distribution of denoised latents from diffusion, this parallel operation of denoising and optimization effectively reduces detail distortions, enhancing face quality.

Furthermore, we prove that the solving process of the HJB equation~\cite{bardi1997optimal, peng1992stochastic, chen2023generative} can be integrated with the diffusion denoising process, as demonstrated below.
The basic HJB Equation can be described as:
\begin{equation}\small
\label{eq:basic_HJB}
\begin{aligned}
     \frac{\partial \mathtt{V}(\bm{x}, t)}{\partial t}+\mathtt{max}_{c}[\mathtt{f}(\bm{x}, \bm{c})+\frac{\partial \mathtt{V}(\bm{x}, t)}{\partial \bm{x}}\cdot\mathtt{g}(\bm{x}, \bm{c})]=0,
\end{aligned}
\end{equation}
where $\mathtt{V}(\bm{x}, t)$ refers to the value function, representing the minimum cost from state $\bm{x}$ at time $t$. $\mathtt{f}(\bm{x}, \bm{c})$ is the immediate cost under the condition $\bm{c}$ in state $\bm{x}$. $\mathtt{g}(\cdot)$ depicts the system dynamics. 
In our settings, the condition $\bm{c}$ indicates the face-aware variable.
Following the previous work~\cite{chen2023generative}, the solving process is formulated as:
\begin{equation}\small
\label{eq:particular_HJB}
\begin{aligned}
     \mathtt{min}_{\bm{c}_{t}}\int_{0}^{1}\frac{1}{2}\left \|\bm{c}_{t}\right \|_{2}^{2}dt+\frac{\bm{r}}{2}\left \|\bm{X}_{1}-\bm{x}_{1}\right \|_{2}^{2}, \bm{X}_{1}\sim \bm{p}_{data}, 
\end{aligned}
\end{equation}
s.t. $d\bm{X}_{t}=\bm{c}_{t}dt$ and $\bm{X}_{0}=\bm{x}_{0}$ (Gaussian noise). $\bm{r}$ is the terminal cost coefficient. 
In our work, we normalize denoising timesteps ${t}'$ (from $\bm{T}$ to $0$) to $[0, 1]$ and set $t=1-{t}'$.
$\bm{T}$ is the maximum denoising timestep. $\bm{X}_{t}$ and $\bm{x}_{t}$ refer to the groundtruth sample and the predicted sample by the model. 
Thus, $\bm{x}_{\text{pred}}$ in Algorithm \ref{alg:face_optimization} is equivalent to $\bm{x}_{1}$.
Following the Pontryagin Maximum Principle~\cite{kirk2004optimal}, we can construct the Hamiltonian equation:
\begin{equation}\small
\label{eq:hamiltonian}
\begin{aligned}
     \mathtt{H}(t,\bm{X},\bm{c}_{t},\bm{\gamma})=-\frac{1}{2}\left \|\bm{c}_{t}\right \|_{2}^{2}+\bm{\gamma}\bm{c}_{t},
\end{aligned}
\end{equation}
where $\bm{\gamma}$ refers to a coefficient.
To minimize Eq. \ref{eq:hamiltonian}, we set $\frac{\partial \mathtt{H}}{\partial \bm{c}_{t}}=0$. The optimal Hamiltonian is described as:
\begin{equation}\small
\label{eq:optimized_hamiltonian}
\begin{aligned}
     \mathtt{H}^{*}=\mathtt{H}(t,\bm{X},{\bm{c}}^{*}_{t},\bm{\gamma})=\frac{1}{2}\bm{\gamma}^{2}, \text{where } {\bm{c}}^{*}_{t}=\bm{\gamma}. 
\end{aligned}
\end{equation}
Then we solve the Hamiltonian equation of motion:
\begin{equation}\small
\label{eq:motion}
\begin{aligned}
     \frac{d\bm{X}_{t}}{dt}&=\frac{\partial \mathtt{H}^{*}}{\partial \bm{\gamma}}=\bm{\gamma}, \\
     \frac{d\bm{\gamma}}{dt}&=\frac{\partial \mathtt{H}^{*}}{\partial \bm{X}}=0.
\end{aligned}
\end{equation}
At the final step $t=1$, from Eq. \ref{eq:particular_HJB} and Eq. \ref{eq:hamiltonian}, we can obtain $\bm{\gamma}_{1}=-\bm{r}\cdot(\bm{X}_{1}-\bm{x}_{1})$. From Eq. \ref{eq:motion}, we can see that $\bm{\gamma}$ is a variable independent of $t$, thereby obtaining $\bm{\gamma}=\bm{\gamma}_{1}=-\bm{r}\cdot(\bm{X}_{1}-\bm{x}_{1})$. We can also get $\bm{X}_{t}=\bm{X}_{0}+\bm{\gamma}t$ $\rightarrow$ $\bm{X}_{1}=\bm{X}_{0}+\bm{\gamma}$ and $\bm{X}_{0}=\bm{X}_{t}-\bm{\gamma}t$. We then obtain ${\bm{c}}^{*}_{t}$:
\begin{equation}\small
\label{eq:optimal_c}
\begin{aligned}
    &\bm{X}_{1}=\bm{X}_{0}+\bm{\gamma}=\bm{X}_{t}-\bm{\gamma}t+\bm{\gamma} \\ 
     \rightarrow&\quad \bm{\gamma}=-\bm{r}\cdot(\bm{X}_{1}-\bm{x}_{1})=-\bm{r}\cdot(\bm{X}_{t} - \bm{\gamma}t + \bm{\gamma} -\bm{x}_{1}), \\
     \rightarrow&\quad {\bm{c}}^{*}_{t}=\bm{\gamma}=\frac{\bm{r}(\bm{x}_{1}-\bm{X}_{t})}{1+\bm{r}(1-t)}.
\end{aligned}
\end{equation}
When $\bm{r} \to \infty$, following Eq. \ref{eq:particular_HJB} ($d\bm{X}_{t}=\bm{c}_{t}dt$) and certainty equivalence~\cite{fleming2012deterministic,chen2023generative} (the stochastic case), we have
\begin{equation}\small
\label{eq:extreme_formula}
\begin{aligned}
     d\bm{X}_{t}=\frac{\bm{x}_{1}-\bm{X}_{t}}{1-t}dt+d\bm{w}_{t},
\end{aligned}
\end{equation}
where $\bm{w}_{t}$ is Brownian motion~\cite{chen2023generative}.
According to EDM~\cite{karras2022elucidating} in SVD~\cite{blattmann2023stable}, where $\bm{X}_{{t}'}=\bm{X}_{data}+{t}'\bm{\varepsilon}$ and $\bm{X}_{data}\sim \bm{p}_{data}$, the current state $\bm{X}_{{t}'}$ is converted to $\bm{X}_{t}=\bm{X}_{1}+(1-t)\bm{\varepsilon}$ in our settings. We use the following Tweedie’s formula~\cite{efron2011tweedie} 
\begin{equation}\small
\label{eq:tweedie}
\begin{aligned}
     \mathtt{E}[\bm{\theta}|\bm{x}]=\bm{x}+\bm{\sigma}^{2}\cdot\nabla\log\mathtt{p}(\bm{x}),
\end{aligned}
\end{equation}
where $\bm{x}|\bm{\theta}\sim\mathcal{N}(\bm{\theta}, \bm{\sigma}^{2})$ and $\mathtt{p}(\cdot)$ is the marginal density of $\bm{x}$, to reform $\bm{X}_{1}$:
\begin{equation}\small
\label{eq:approximation}
\begin{aligned}
     \bm{X}_{1}=\mathtt{E}[\bm{X}_{1}|\bm{X}_{t}]=\bm{X}_{t}+(1-t)^{2}\nabla\log\mathtt{p}(\bm{X}_{t}).
\end{aligned}
\end{equation}
$\bm{x}_{1}$ aims to approximate $\bm{X}_{1}$. Thus, we substitute Eq. \ref{eq:approximation} in Eq. \ref{eq:extreme_formula} for obtaining the ultimate formula:
\begin{equation}\small
\label{eq:final}
\begin{aligned}
     d\bm{X}_{t}&=\frac{\bm{X}_{t}+(1-t)^{2}\nabla\log\mathtt{p}(\bm{X}_{t})-\bm{X}_{t}}{1-t}dt+d\bm{w}_{t} \\
     &=(1-t)\cdot\nabla\log\mathtt{p}(\bm{X}_{t})dt+d\bm{w}_{t}.
\end{aligned}
\end{equation}

It is evident that Eq. \ref{eq:final} and SDE formulation~\cite{song2020score} are structurally the same, thus we can seamlessly incorporate the solution process of the HJB equation into the diffusion denoising for ID preservation.

\subsection{Training}
\label{sec: training}
As illustrated in Fig. \ref{fig:framework}, we use the reconstruction loss to train our model, with trainable components including a U-Net, a FaceEncoder, and a PoseNet. 
We introduce face masks $\bm{M}$, extracted by ArcFace~\cite{deng2019arcface} from the input video frames to enhance the modeling of face regions:
\begin{equation}\small
\label{eq:loss}
\begin{aligned}
     \mathcal{L}=\mathbb{E}_{\varepsilon}(\left \| (\bm{z}_{gt}-\bm{z}_{\varepsilon})\odot  (1+\bm{M})  \right \|^{2}),
\end{aligned}
\end{equation}
where $\bm{z}_{gt}$ and $\bm{z}_{\varepsilon}$ are diffusion latents and denoised latents.

\begin{figure}[t!]
\begin{center}
\includegraphics[width=1\linewidth]{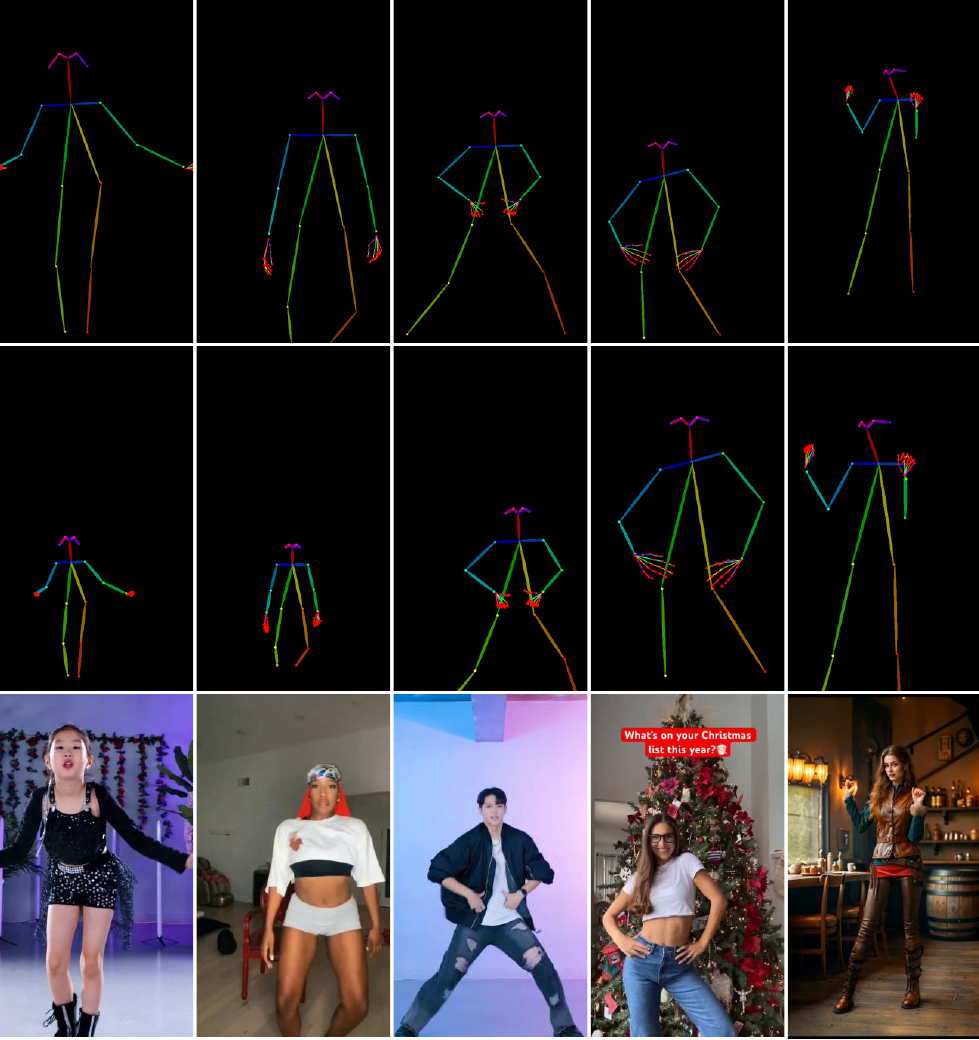}
\end{center}
\vspace{-0.5cm}
   \caption{Examples from MisAlign100. The first row, the second row, and the third row refer to the original driven poses, modified driven poses, and corresponding reference image, respectively.
   }
\label{fig:misalign100}
\vspace{-0.5cm}
\end{figure}
 
\section{Experiments}
\subsection{Implementation Details}
As previous works do not open-source their training datasets, we collect 5K videos (60-90 seconds long) from the internet to train our model. We use DWPose~\cite{yang2023effective} to extract skeleton poses. 
Following~\cite{wang2024disco, hu2024animate, xu2024magicanimate, zhu2024champ, wang2024unianimate, zhang2024mimicmotion, peng2024controlnext, tan2024animate_x}, we evaluate our model on TikTok dataset~\cite{jafarian2021learning}.
We also select 100 unseen videos (the MisAlign100 dataset) from the internet, featuring scenarios with significant misalignment. Following recent works~\cite{zhang2024mimicmotion, peng2024controlnext}, the U-Net uses pre-trained weights of Stable Video Diffusion~\cite{blattmann2023stable}, while the PoseNet, Face Encoder, and alignment block are trained from scratch. 
Regarding the Transformer Encoders ($\mathtt{Encoder}_{m}(\cdot)$ and $\mathtt{Encoder}_{svd}(\cdot)$ in our learnable alignment, they all share the same architecture, comprising two modules, each containing a self-attention block and an FFN. Notably, since $\mathtt{Encoder}_{m}(\cdot)$ and $\mathtt{Encoder}_{svd}(\cdot)$ both model keypoint sequences, where each token corresponds to a skeleton node, we apply position embeddings to the input sequences before passing them to the self-attention layers of the encoders.
Our ID-Adapter uses pre-trained weights of spatial cross-attention blocks in Stable Video Diffusion. 
Our model is trained for 20 epochs on 8 NVIDIA A100 80G GPUs, with a batch size of 1 per GPU. The learning rate is set to 1$e$-5.
Our HJB-based face optimization is applied exclusively during the first 10 denoising steps at inference.

\begin{figure*}[t!]
\begin{center}
\includegraphics[width=0.98\linewidth]{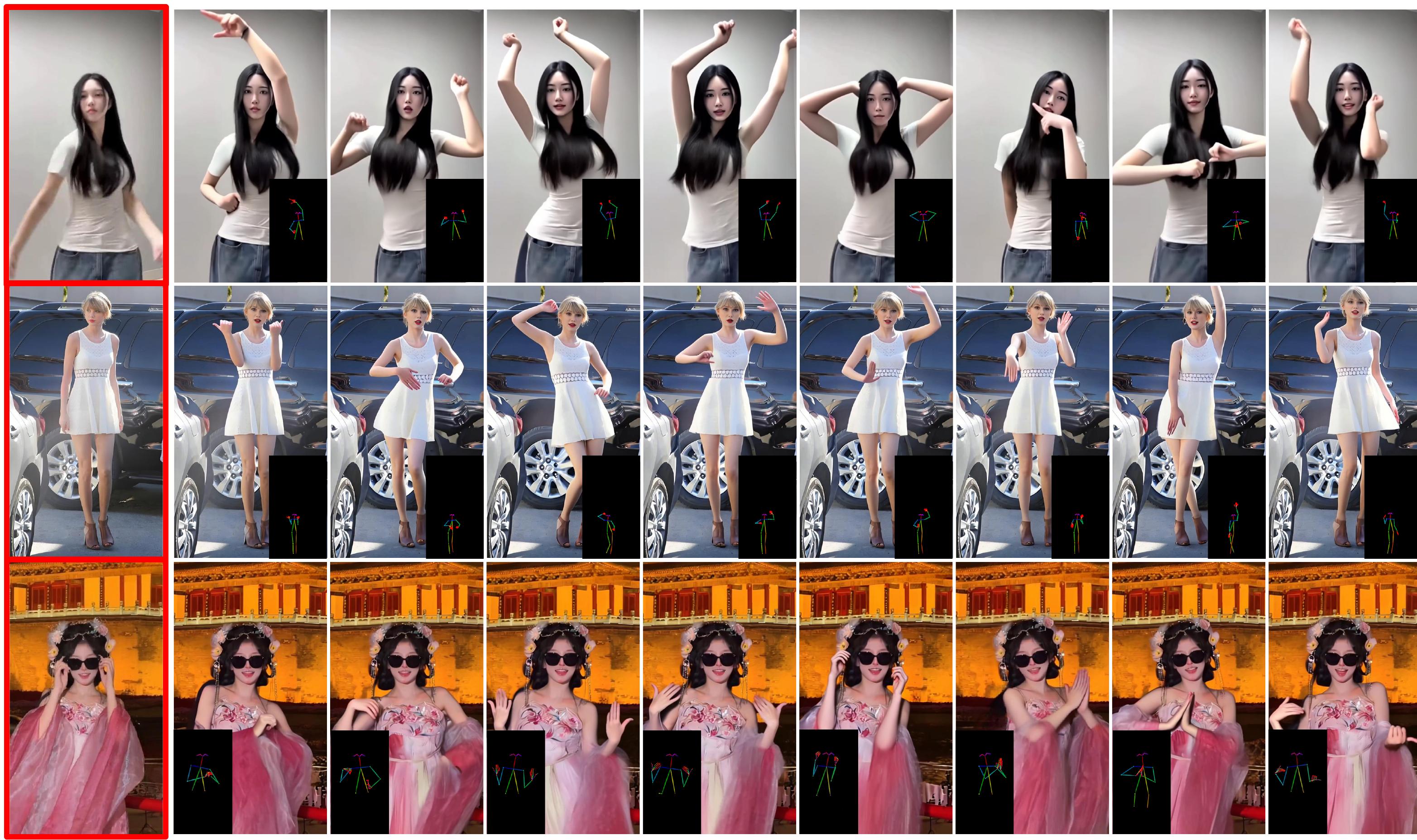}
\end{center}
\vspace{-0.55cm}
   \caption{Animation results generated by StableAnimator++. The images with red borders are the reference images. The presented pose skeletons are dramatically misaligned with the reference image in body size and position.
   }
\label{fig:animation_results}
\vspace{-0.45cm}
\end{figure*}

\begin{table*}[t!]\small
\caption{Quantitative comparisons on TikTok dataset and MisAlign100.
}
\begin{center}
\renewcommand\arraystretch{1.1}
\scalebox{0.9}{
\begin{tabular}{l|cccccc|c|c}
\toprule
Model                & L1(E-4)$\downarrow$   & PSNR~\cite{hore2010image}$\uparrow$        & PSNR*~\cite{wang2024disco}$\uparrow$       & SSIM$\uparrow$        & LPIPS$\downarrow$       & CSIM~\cite{guo2024liveportrait}$\uparrow$       & FVD$\downarrow$            & Mem$\downarrow$    \\ \midrule
MRAA~\cite{siarohin2021motion}                 & 3.21/3.88 & -/18.12     & 18.14/9.81  & 0.672/0.285 & 0.296/0.637 & 0.248/0.163 & 284.82/1782.57 & \textbf{5.4G}   \\
DisCo~\cite{wang2024disco}                & 3.78/3.84 & 29.03/18.58 & 16.55/9.84  & 0.668/0.293 & 0.292/0.634 & 0.315/0.202 & 292.80/1745.13 & 18.7G  \\
MagicAnimate~\cite{xu2024magicanimate}         & 3.13/3.32 & 29.16/18.94 & -/10.06     & 0.714/0.315 & 0.239/0.623 & 0.462/0.268 & 179.07/1342.66 & 20.84G \\
AnimateAnyone~\cite{hu2024animate}        & -/3.27    & 29.56/19.28 & -/10.16     & 0.718/0.324 & 0.285/0.619 & 0.457/0.261 & 171.90/1287.42 & 11.18G \\
Champ~\cite{zhu2024champ}                & 2.94/3.04 & 29.91/22.88 & -/12.17     & 0.802/0.389 & 0.234/0.522 & 0.350/0.307 & 160.82/1046.48 & 13.20G \\
Unianimate~\cite{wang2024unianimate}           & \textbf{2.66}/2.87 & 30.77/25.85 & 20.58/14.52 & 0.811/0.467 & 0.231/0.465 & 0.479/0.324 & 148.06/768.05  & 6.11G  \\
MimicMotion~\cite{zhang2024mimicmotion}          & 5.85/3.80 & -/17.73     & 14.44/9.88  & 0.601/0.298 & 0.414/0.628 & 0.262/0.245 & 232.95/1652.78 & 8.60G  \\
ControlNeXt~\cite{peng2024controlnext}          & 6.20/2.92 & -/24.69     & 13.83/13.41 & 0.615/0.482 & 0.416/0.516 & 0.360/0.278 & 326.57/687.34  & 12.23G \\
Animate-X~\cite{tan2024animate_x}            & 2.70/2.83 & 30.78/26.82 & 20.77/16.38 & 0.806/0.512 & 0.232/0.429 & 0.475/0.391 & 139.01/675.26  & 14.3G  \\ \midrule
StableAnimator++ (Ours) & 2.90/\textbf{2.74} & \textbf{30.81}/\textbf{30.17} & \textbf{20.79}/\textbf{18.22} & \textbf{0.816}/\textbf{0.709} & \textbf{0.230}/\textbf{0.375} & \textbf{0.831}/\textbf{0.802} & \textbf{122.47}/\textbf{384.27}  & 11.40G \\ \bottomrule
\end{tabular}
}
\end{center}
\vspace{0pt}
\makebox[\linewidth]{%
  \begin{minipage}{0.9\linewidth}  
    \scriptsize\justifying\noindent Mem refers to GPU memory when manipulating 16 frames ($576\times1024$). 
      In the table elements $a$ / $b$, $a$, and $b$ refer to the result on the TikTok dataset 
      and MisAlign100, respectively. We reference competitors' results on the TikTok dataset 
      from their papers, with $-$ indicating missing reports.
  \end{minipage}%
}
\label{table:quantitative_comparisons}
\vspace{-0.20in}
\end{table*}

\subsection{Data Collection}
We collect our training videos from YouTube and TikTok. The raw videos are fed to the InsightFace~\cite{deng2019arcface} and Cotracker~\cite{karaev23cotracker} models to filter out those with low facial quality or significant camera motion (such as shot changes or background variance). 
We further apply DWPose to remove any videos where the skeletons lack more than 70\% keypoints, thus obtaining our dataset.

Regarding the MisAlign100 dataset, we collect 100 unseen videos (10-20 seconds long) from the internet to construct the testing dataset MisAlign100. Some cases are shown in Fig. \ref{fig:misalign100}.
The videos originate from various social media platforms, including YouTube, TikTok, and BiliBili, featuring individuals of diverse backgrounds and genders. They are captured in full-body, half-body, and close-up shots across a range of indoor and outdoor environments.
In contrast to the existing open-source animation testing dataset (TikTok dataset), our MisAlign100 involves more complex motion patterns and intricate appearance information. Each driven pose is randomly rotated, scaled, and translated to simulate the misalignment which is commonly encountered in real-world scenarios, making it more challenging to maintain ID consistency. 

\begin{figure*}[t!]
\begin{center}
\includegraphics[width=0.98\linewidth]{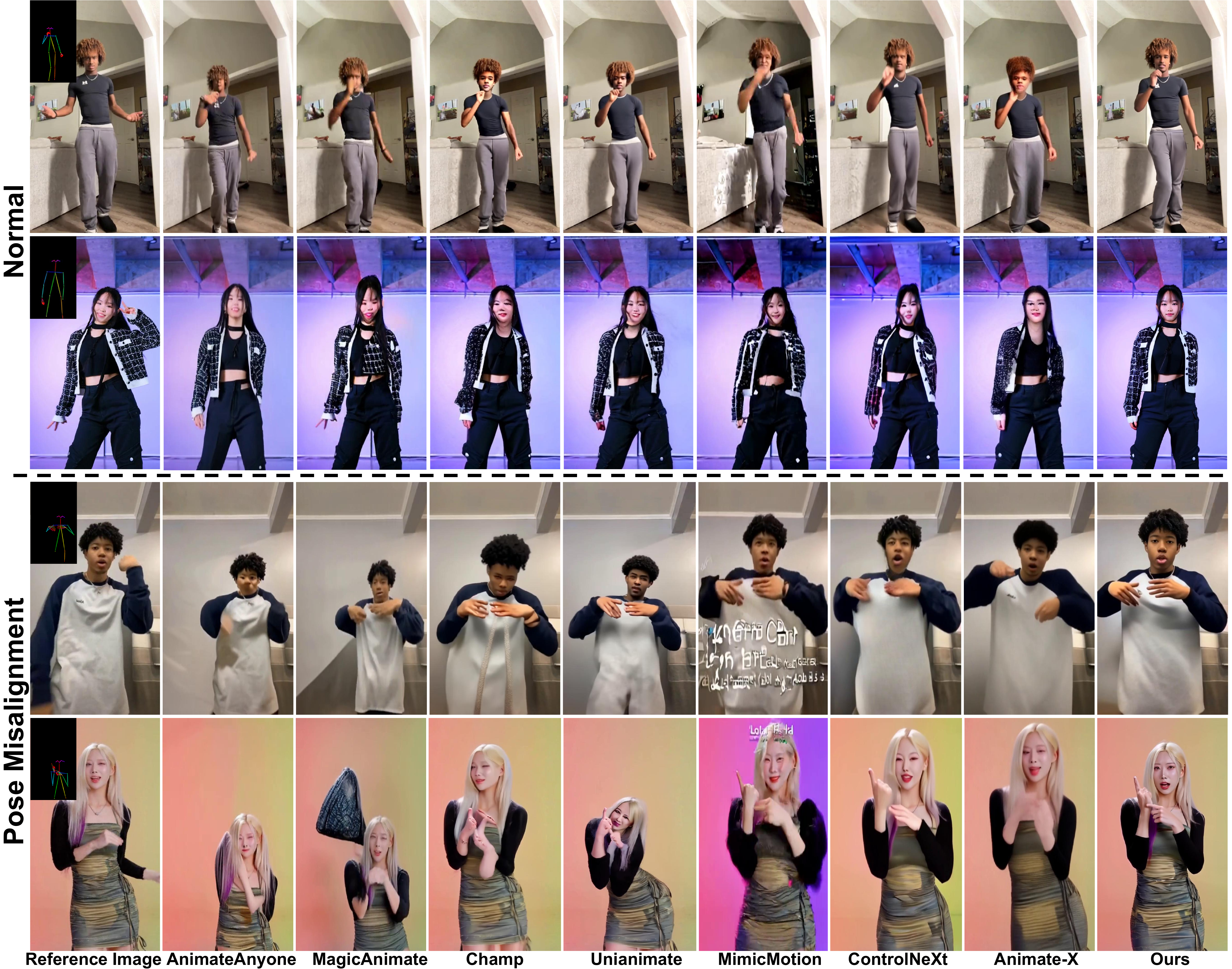}
\end{center}
\vspace{-0.55cm}
   \caption{Qualitative comparisons with state-of-the-art methods. The skeletons in the third and fourth rows are misaligned with the reference image in terms of body size or position.
   More examples can be found in the supplementary material.}
\label{fig:comparison}
\vspace{-0.45cm}
\end{figure*}

\subsection{Animation Results}
We demonstrate the animation results in Fig. \ref{fig:animation_results}. 
We can observe that our StableAnimator++ can perform a wide range of human image animation while simultaneously preserving the reference consistency, including the protagonist's appearance details and the background layouts. 
Each case involves a protagonist with complex appearance and intricate motion dynamics, while the reference image and driven video exhibit significant discrepancies in body shape and position.
More cases are shown in the Sec.VIII of the Supp.

\subsection{Comparison with State-of-the-Art Methods}
\textbf{Quantitative results.}
We compare with recent human image animation models, including GAN-based models (MRAA~\cite{siarohin2021motion}) and diffusion models (AnimateAnyone~\cite{hu2024animate}, MagicAnimate~\cite{xu2024magicanimate}, Champ~\cite{zhu2024champ}, Unianimate~\cite{wang2024unianimate}, MimicMotion~\cite{zhang2024mimicmotion}, ControlNeXt~\cite{peng2024controlnext}, Animate-X~\cite{tan2024animate_x}), as shown in Table \ref{table:quantitative_comparisons}.
CSIM~\cite{guo2024liveportrait} evaluates the cosine similarity between the facial embeddings of two images.
Based on previous studies that assess quantitative results using the self-driven and reconstruction approach, we perform quantitative comparisons with the above competitors on the TikTok dataset~\cite{jafarian2021learning} and MisAlign100. Notably, we randomly scale / translate / rotate the driven poses before evaluating on MisAlign100 to simulate misalignment. All competitors are trained on our dataset before evaluating on MisAlign100 to ensure a fair comparison. Since AnimateAnyone lacks a default alignment operation, we apply ControlNeXt's alignment to it.
We can see that our StableAnimator++ outperforms all competitors on MisAlign100 in both video fidelity and single-frame quality under significant misalignment scenarios while achieving relatively promising performance on the TikTok dataset.
In particular, StableAnimator++ outperforms the leading competitor, Animate-X, by 35.6\% and 41.1\% in CSIM across two datasets, without sacrificing video fidelity and single-frame quality.

\begin{table}[t!]\small
\caption{Ablation study on core components. 
}
\vspace{-0.15in}
\begin{center}
\renewcommand\arraystretch{1.1}
\scalebox{0.70}{
\begin{tabular}{lccccc|c}
\toprule
Model            & L1$\downarrow$                                   & PSNR$\uparrow$                               & SSIM$\uparrow$                               & LPIPS$\downarrow$                              & CSIM$\uparrow$                                & FVD$\downarrow$                                 \\ \midrule
\textit{w/o} Pose Align (SA\cite{tu2024stableanimator})       & 3.58E-4                              & 18.51                              & 0.298                              & 0.630                              & 0.448                               & 1635.24                              \\
\textit{w/o} Prediction   & 2.82E-4                              & 26.84                              & 0.542                              & 0.424                              & 0.726                               & 552.13                              \\
\textit{w/o} Face Masks   & \multicolumn{1}{l}{2.79E-4}          & \multicolumn{1}{l}{27.11}          & \multicolumn{1}{l}{0.653}          & \multicolumn{1}{l}{0.386}          & \multicolumn{1}{l|}{0.694}          & \multicolumn{1}{l}{458.91}          \\
\textit{w/o} Face Encoder & \multicolumn{1}{l}{2.82E-4}          & \multicolumn{1}{l}{27.03}          & \multicolumn{1}{l}{0.647}          & \multicolumn{1}{l}{0.390}          & \multicolumn{1}{l|}{0.572}          & \multicolumn{1}{l}{441.16}          \\
\textit{w/o} Distribution Align   & 2.85E-4                              & 25.98                              & 0.496                              & 0.435                              & 0.707                               & 587.36                              \\
\textit{w/o} Optimization & 2.78E-4                              & 27.72                              & 0.685                              & 0.382                              & 0.778                               & 404.28                              \\ \midrule
Ours             & \multicolumn{1}{l}{\textbf{2.74E-4}} & \multicolumn{1}{l}{\textbf{30.17}} & \multicolumn{1}{l}{\textbf{0.709}} & \multicolumn{1}{l}{\textbf{0.375}} & \multicolumn{1}{l|}{\textbf{0.802}} & \multicolumn{1}{l}{\textbf{384.27}} \\ \bottomrule
\end{tabular}
}
\end{center}
\vspace{0pt}
\makebox[\linewidth]{%
  \begin{minipage}{0.95\linewidth}  
    \scriptsize\justifying\noindent  \textit{w/o} Prediction removes learnable layers in our pose alignment, directly applying SVD outputs to align poses. Face Masks and Distribution Align refer to face masks in the loss and distribution alignment of our ID Adapter. SA refers to StableAnimator~\cite{tu2024stableanimator}.
  \end{minipage}%
}
\label{table:ablation_core}
\vspace{-0.15in}
\end{table}

\noindent\textbf{Qualitative Results.}
The qualitative results are shown in Fig. \ref{fig:comparison}. All qualitative results in the paper are in the cross-ID setting~\cite{zhu2024champ}.
MagicAnimate~\cite{xu2024magicanimate}, AnimateAnyone~\cite{hu2024animate}, and Champ~\cite{zhu2024champ} exhibit face / body distortion and clothing changes, while Unianimate~\cite{wang2024unianimate} and Animate-X~\cite{tan2024animate_x} accurately modify the reference motion, and MimicMotion~\cite{zhang2024mimicmotion} and ControlNeXt~\cite{peng2024controlnext} effectively preserve clothing details.
However, all competitors still struggle with face/body distortion and blurry noises in both normal and pose-misaligned scenarios. 
In contrast, our StableAnimator++ accurately animates images based on the given pose sequences while preserving reference identities even in misalignment scenarios, showcasing the superiority of our model in identity retention and in generating precise, vivid animations.

\begin{figure}[t!]
\begin{center}
\includegraphics[width=1\linewidth]{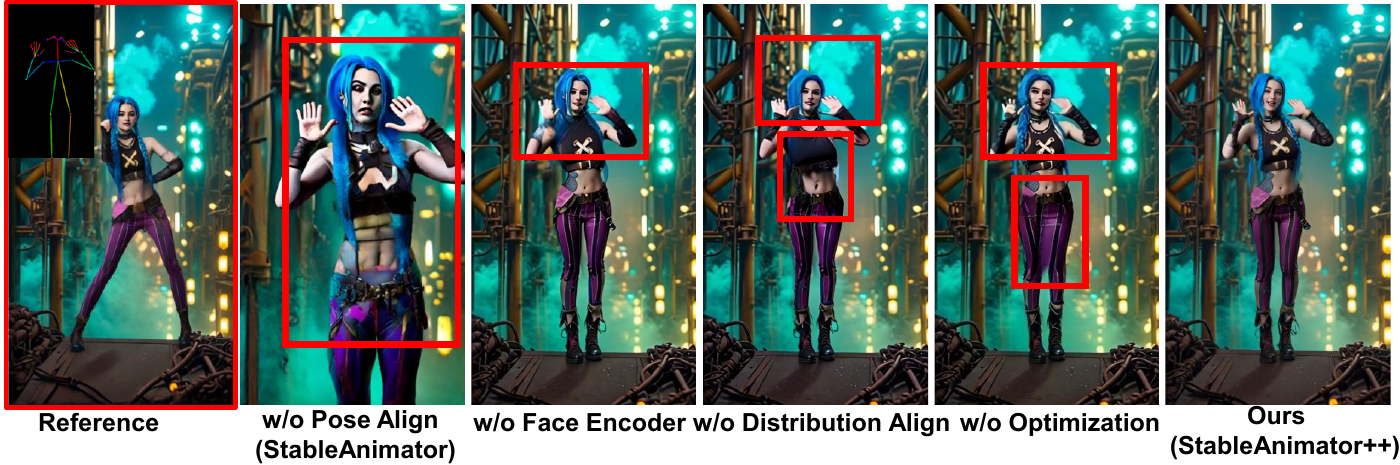}
\end{center}
\vspace{-0.6cm}
   \caption{Ablations on core components of StableAnimator++. 
   The presented skeleton is misaligned with the reference image in body size and position.
   }
\label{fig:ablation_core}
\vspace{-0.40cm}
\end{figure}

\subsection{Ablation Study}
\noindent \textbf{Pose Alignment.}
We conduct an ablation study to validate the contributions of core components in StableAnimator++, as shown in Fig. \ref{fig:ablation_core} and Table \ref{table:ablation_core}. All quantitative ablation studies are conducted on the MisAlign100 dataset. 
The \textit{w/o} Pose Align setting is equivalent to StableAnimator (SA)~\cite{tu2024stableanimator}.
We can observe that removing core components dramatically deteriorates performance, particularly in face-related regions (CSIM), indicating that each core component can promote both single-frame quality and video fidelity while maintaining identity consistency even in misalignment scenarios.

\begin{figure}[t!]
\begin{center}
\includegraphics[width=1\linewidth]{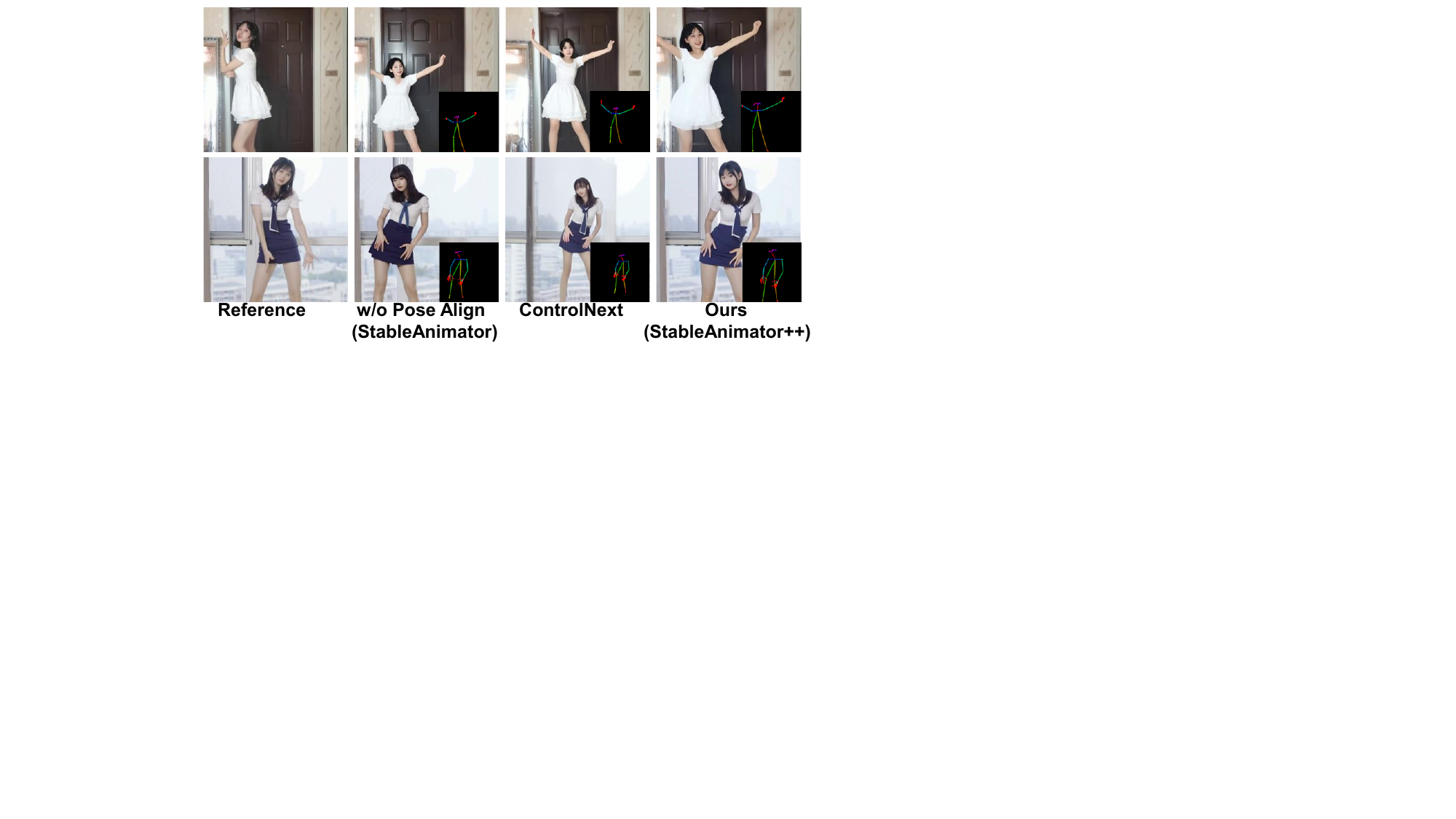}
\end{center}
\vspace{-0.4cm}
   \caption{Ablations on the alignment. The poses in the last two columns are aligned by the respective methods.
   }
\label{fig:ablation_alignment}
\vspace{-0.40cm}
\end{figure}

\begin{figure}[t!]
\begin{center}
\includegraphics[width=1\linewidth]{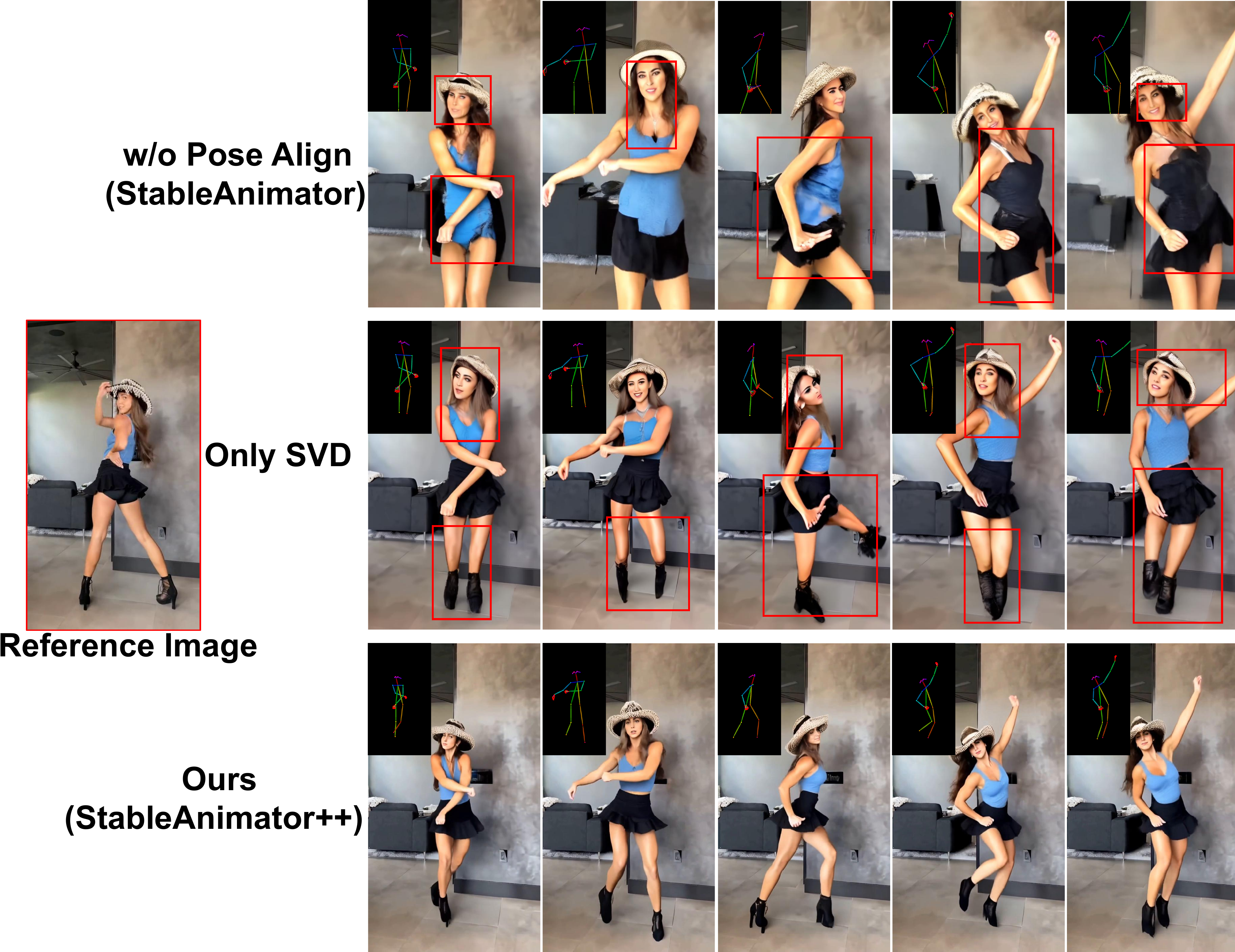}
\end{center}
\vspace{-0.4cm}
   \caption{Ablations on the alignment. The poses in the last two rows are aligned by the respective methods. Only SVD removes learnable layers in our alignment, directly applying SVD outputs to align poses. 
   }
\label{fig:ablation_alignment_svd}
\vspace{-0.40cm}
\end{figure}

\begin{figure}[t!]
\begin{center}
\includegraphics[width=1\linewidth]{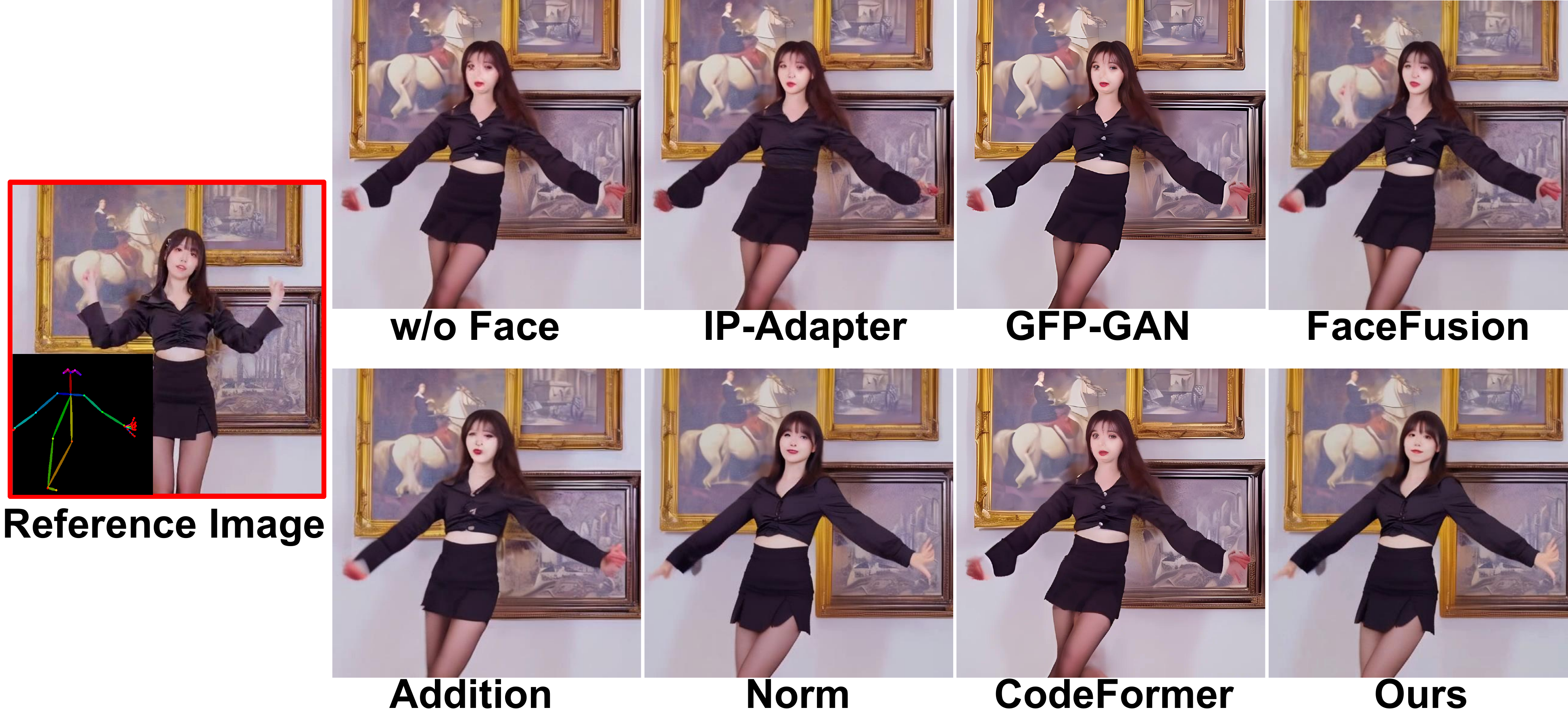}
\end{center}
\vspace{-0.6cm}
   \caption{Ablation study on face enhancement strategies.
   }
\label{fig:ablation_face}
\vspace{-0.35cm}
\end{figure}

\begin{table}[t!]\small
\caption{Ablation study on the alignment.}
\begin{center}
\renewcommand\arraystretch{1.1}
\scalebox{0.70}{
\begin{tabular}{lccccccc}
\toprule
Model          & L1$\downarrow$               & PSNR$\uparrow$           & SSIM$\uparrow$           & CSIM$\uparrow$          & Dis$\downarrow$   & FVD$\downarrow$             \\ \midrule
\textit{w/o} Pose Alignment (SA\cite{tu2024stableanimator})  & 3.58E-4          & 18.51          & 0.298          & 0.448          & 0.597 & 1635.24         \\
\textit{w/} ControlNeXt & 2.88E-4          & 25.32          & 0.488          & 0.661          & 0.430 & 625.97          \\
\textit{w/o} Prediction & 2.84E-4          & 26.94          & 0.510          & 0.702          & 0.345 & 571.56          \\ \midrule
Ours           & \textbf{2.74E-4} & \textbf{30.17} & \textbf{0.709} & \textbf{0.802} & \textbf{0.105} & \textbf{384.27} \\ \bottomrule
\end{tabular}
}
\end{center}
\vspace{0pt}
\makebox[\linewidth]{%
  \begin{minipage}{0.95\linewidth}  
    \scriptsize\justifying\noindent  Dis is the average Euclidean distance between aligned poses and ground-truths.
  \end{minipage}%
}
\label{table:ablation_alignment}
\vspace{-0.15in}
\end{table}

We further compare our alignment with the current keypoint alignment approach~\cite{peng2024controlnext}, as shown in Fig. \ref{fig:ablation_alignment} and Table \ref{table:ablation_alignment}. We replace our alignment with ControlNeXt's alignment, which is also commonly used in current animation models~\cite{zhang2024mimicmotion, wang2024unianimate}.
Fig. \ref{fig:ablation_alignment_svd} ablates the effectiveness of the SVD-based transformation. 
By analyzing the results, we can gain the following observations: 
(1) StableAnimator~\cite{tu2024stableanimator} exhibits noticeable face/body distortions in scenarios with significant pose misalignment between the reference and the driving video.
(2) ControlNeXt's alignment reduces body distortion but degrades video fidelity and reference consistency, as its aligned driven poses fail to match the reference image in body size and position, creating a conflict between appearance preservation and motion modeling.
(3) Directly using SVD outputs for alignment enhances single-frame quality but compromises reference consistency. The plausible reason is that the transformation matrices of SVD are not particularly accurate, leading to a loss of semantic details. 
(4) StableAnimator++ can effectively preserve identity while achieving high video fidelity, as our pose alignment can dramatically reduce the gap between the reference and driven poses. More ablation studies are in Sec. IV of the Supp.

\noindent \textbf{Face Enhancement Strategies.}
We conduct an ablation study regarding current face enhancement approaches, as shown in Table \ref{table:ablation_face} and Fig. \ref{fig:ablation_face}. We replace our face-related components with the commonly used IP-Adapter and FaceFusion. 
\textbf{We temporarily apply our pose alignment to the MisAlign100 dataset to obtain aligned poses for a fair comparison in the following ablation studies.}
By analyzing the results, we can gain the following observations:
(1) IP-Adapter can improve the ID consistency, while the video fidelity and single-frame quality dramatically degrade. The plausible reason is that directly inserting the IP-Adapter hinders its ability to adapt to spatial representation distribution variations during temporal modeling, thereby deteriorating the capacity of the video diffusion model. (2) The third-party post-processing face-swapping tool FaceFusion refines the face quality but relatively degrades the video fidelity. The underlying reason is that the third-party post-processing operates in a different domain from the diffusion model, leading to a loss of semantic details and disrupting video fidelity. (3) StableAnimator++ can significantly refine the face quality while maintaining high video fidelity since our model remains in the same domain as the video diffusion model due to the distribution-aware end-to-end pipeline.

We further conduct a comparison between our StableAnimator++ and other facial restoration models (GFP-GAN~\cite{wang2021gfpgan} and CodeFormer~\cite{zhou2022codeformer}), as shown in Fig. \ref{fig:ablation_face}. It is noticeable that our StableAnimator++ has the best identity-preserving capability compared with other competitors, demonstrating the superiority of our StableAnimator++ regarding identity consistency. By contrast, GFP-GAN and CodeFormer suffer from serious facial distortion and over-sharpening. The plausible reason is that \textit{w/o} Face cannot synthesize the precise facial layout, which in turn undermines the effectiveness of subsequent facial restoration processes. This represents a fundamental limitation of post-processing-based face enhancement strategies.

\begin{table}[t!]\small
\caption{Ablation study on face enhancement methods.
}
\vspace{-0.15in}
\begin{center}
\renewcommand\arraystretch{1.1}
\scalebox{0.8}{
\begin{tabular}{lccccc|c}
\toprule
Model      & L1$\downarrow$                          & PSNR$\uparrow$                      & SSIM$\uparrow$                      & LPIPS$\downarrow$                     & CSIM$\uparrow$           & FVD$\downarrow$                        \\ \midrule
\textit{w/o} Face   & 2.83E-4                     & 26.75                     & 0.741                     & 0.264                     & 0.324          & 371.38                     \\
IP-Adapter~\cite{ye2023ip-adapter} & 3.88E-4                     & 18.86                     & 0.672                     & 0.287                     & 0.511          & 484.77                     \\
FaceFusion~\cite{facefusion} & 3.31E-4 & 23.05 & 0.734 & 0.265 & 0.798          & 405.16 \\ \midrule
Ours       & \textbf{2.71E-4}            & \textbf{28.85}            & \textbf{0.784}            & \textbf{0.223}            & \textbf{0.805} & \textbf{349.94}            \\ \bottomrule
\end{tabular}
}
\end{center}
\vspace{0pt}
\makebox[\linewidth]{%
  \begin{minipage}{0.95\linewidth}  
    \scriptsize\justifying\noindent  \textit{w/o} Face refers to the exclusion of any face-related strategies.
  \end{minipage}%
}
\label{table:ablation_face}
\vspace{-0.25in}
\end{table}

\begin{table}[t!]\small
\caption{Ablation study on the distribution-based alignment.
}
\vspace{-0.2in}
\begin{center}
\renewcommand\arraystretch{1.1}
\scalebox{0.8}{
\begin{tabular}{lccccc|c}
\toprule
Model                 & L1$\downarrow$               & PSNR$\uparrow$           & SSIM$\uparrow$           & LPIPS$\downarrow$          & CSIM$\uparrow$           & FVD$\downarrow$             \\ \midrule
Addition & 3.11E-4          & 23.45          & 0.713          & 0.276          & 0.716          & 412.52          \\
Norm         & 2.73E-4          & 26.67          & 0.758          & 0.257          & 0.776          & 382.49          \\ \midrule
Ours                  & \textbf{2.71E-4} & \textbf{28.85} & \textbf{0.784} & \textbf{0.223} & \textbf{0.805} & \textbf{349.94} \\ \bottomrule
\end{tabular}
}
\end{center}
\vspace{0pt}
\makebox[\linewidth]{%
  \begin{minipage}{0.95\linewidth}  
    \scriptsize\justifying\noindent  Addition and Norm refer to element-wise addition and normalization. 
  \end{minipage}%
}
\label{table:ablation_distribution_alignment}
\vspace{-0.2in}
\end{table}

\begin{table}[t!]\small
\caption{Ablation study on the optimization.
}
\vspace{-0.15in}
\begin{center}
\renewcommand\arraystretch{1.1}
\scalebox{0.75}{
\begin{tabular}{lccccc|c}
\toprule
Model                                    & L1$\downarrow$               & PSNR$\uparrow$           & SSIM$\uparrow$           & LPIPS$\downarrow$          & CSIM$\uparrow$           & FVD$\downarrow$             \\ \midrule
Magic+IP                          & 3.85E-4          & 23.14          & 0.689          & 0.286          & 0.541          & 836.33          \\
Magic+FaceFusion                  & 3.31E-4          & 26.42          & 0.725          & 0.268          & 0.796          & 412.40          \\
Magic+Opt                & 3.02E-4          & 27.56          & 0.762          & 0.258          & 0.480          & 381.61          \\
Magic+IP+Opt             & 3.61E-4          & 26.12          & 0.714          & 0.279          & 0.624          & 754.34          \\
Magic+FE+ID              & 2.85E-4          & 27.89          & 0.767          & 0.248          & 0.775          & 376.43          \\
Magic+FE+ID+Opt & \textbf{2.69E-4} & \textbf{28.13} & \textbf{0.775} & \textbf{0.241} & \textbf{0.798} & \textbf{355.23} \\ \bottomrule
\end{tabular}
}
\end{center}
\vspace{0pt}
\makebox[\linewidth]{%
  \begin{minipage}{0.95\linewidth}  
    \scriptsize\justifying\noindent  Magic, IP, ID, FE, and Opt refer to MagicAnimate, IP-Adapter, our ID Adapter, our Face Encoder, and our Optimization, respectively.
  \end{minipage}%
}
\label{table:ablation_opt}
\vspace{-0.3in}
\end{table}

\noindent \textbf{Feature Distortion.}
We conduct a comparison between our distribution alignment in the ID-Adapter and other types of feature injection, as shown in Table \ref{table:ablation_distribution_alignment} and Fig. \ref{fig:ablation_face}. Norm refers to $\bar{\bm{z}}_{i}^{face}$=$\frac{\bm{z}_{i}^{face}-\bm{\mu}_{face}}{\bm{\sigma}_{face}}$. We can see that Addition and Norm fail to eliminate the interference of spatial feature distortion after temporal modeling, thereby achieving suboptimal results. By contrast, our alignment integrates the mean and standard deviation from both cross-attention features, significantly mitigating the impact of feature distortion.

\begin{figure}[t!]
\begin{center}
\includegraphics[width=1\linewidth]{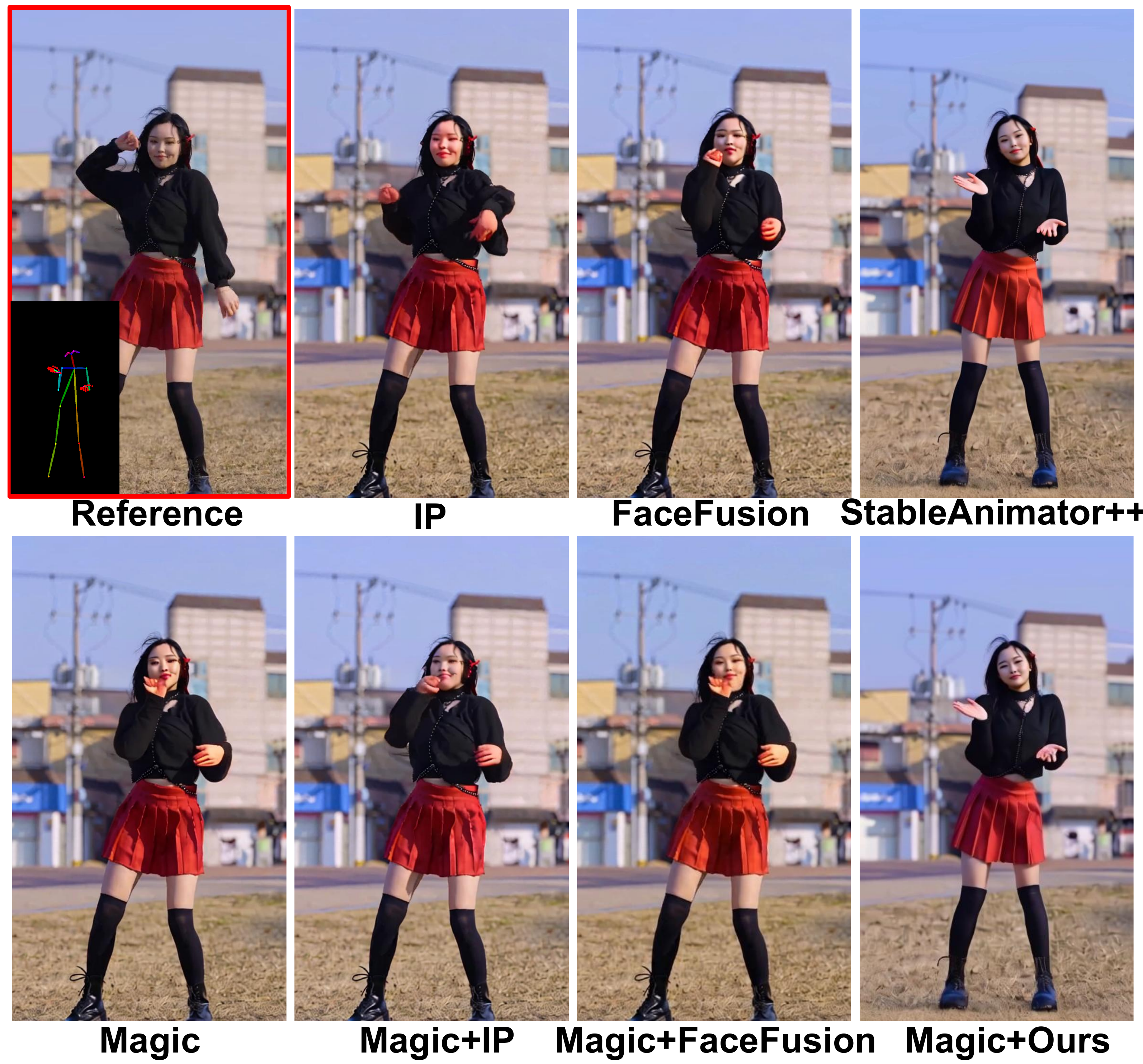}
\end{center}
\vspace{-0.6cm}
   \caption{Ablation study on different backbones.
   }
\label{fig:ablation_opt}
\vspace{-0.5cm}
\end{figure}

\noindent \textbf{Face Optimization.}
To validate the significance of our face optimization strategy, we conduct an ablation regarding different diffusion backbones. The results are in Table \ref{table:ablation_opt} and Fig. \ref{fig:ablation_opt}. MagicAnimate is based on SD~\cite{rombach2022high}+AnimateDiff~\cite{guo2023animatediff}. We have the following observations: (1) Common face enhancement strategies (IP-Adapter and FaceFusion) also degrade the video fidelity and single-frame quality of MagicAnimate, indicating that spatial feature distortion indeed occurs across different diffusion-based backbones. 
(2) Magic+Opt boosts overall performance, showing that our face optimization enhances the diffusion model even without any explicit introduction of face-related adapters. The results of Magic+IP+Opt indicate that our optimization can mitigate the deterioration in fidelity due to the introduction of IP-Adapter while improving face quality to some extent.  
(3) The last two rows of Table \ref{table:ablation_opt} show that our face optimization can still work in different diffusion-based backbones.

\begin{figure}[t!]
\begin{center}
\includegraphics[width=1\linewidth]{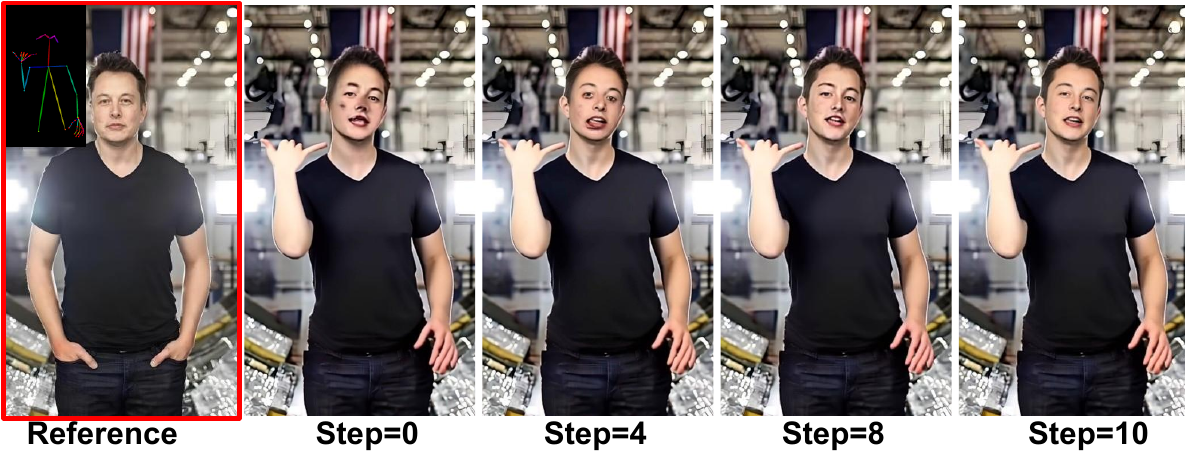}
\end{center}
\vspace{-0.6cm}
   \caption{Visual comparison of HJB-based face optimization at different denoising (optimization) steps.
   }
\label{fig:more_visual_comparison}
\vspace{-0.4cm}
\end{figure}

Fig. \ref{fig:more_visual_comparison} shows a detailed visual comparison, where the step refers to the optimization step in HJB-based optimization. The facial quality progressively improves, which indicates the significance of our face optimization in terms of identity preservation. However, increasing the number of optimization steps introduces higher inference latency, and excessive steps tend to over-sharpen facial details. Thus, we empirically set the total number of steps to 10 as an optimal trade-off between quality and efficiency.

\begin{table}[t!]\small
\caption{Comparison results on inference latency.
}
\vspace{-0.15in}
\begin{center}
\renewcommand\arraystretch{1.1}
\scalebox{0.75}{
\begin{tabular}{lcc|cc}
\toprule
Model         & PSNR$\uparrow$           & FVD$\downarrow$             & Mem$\downarrow$            & Inference Latency$\downarrow$ \\ \toprule
MagicAnimate~\cite{xu2024magicanimate}  & 18.94          & 1342.66         & 20.84G         & 82s               \\
AnimateAnyone~\cite{hu2024animate} & 19.28          & 1287.42         & 11.18G         & 75s               \\
Champ~\cite{zhu2024champ}         & 22.88          & 1046.48         & 13.20G          & 145s              \\
Unianimate~\cite{wang2024unianimate}    & 25.85          & 768.05          & \textbf{6.11G} & 86s               \\
MimicMotion~\cite{zhang2024mimicmotion}   & 17.73          & 1652.78         & 8.60G           & \textbf{60s}      \\
ControlNeXt~\cite{peng2024controlnext}   & 24.69          & 687.34          & 12.23G         & 139s              \\
Animate-X~\cite{tan2024animate_x}     & 26.82          & 575.26           & 14.30G          & 182s              \\ \midrule
Ours          & \textbf{30.17} & \textbf{384.27} & 11.4G          & 84s               \\ \bottomrule
\end{tabular}
}
\end{center}
\label{table:inference_latency}
\vspace{-0.20in}
\end{table}

\begin{table}[t!]\small
\caption{Comparison results on anthropomorphic characters.
}
\vspace{-0.15in}
\begin{center}
\renewcommand\arraystretch{1.1}
\scalebox{0.75}{
\begin{tabular}{lccccc}
\toprule
Model       & L1$\downarrow$               & PSNR*$\uparrow$           & SSIM$\uparrow$           & LPIPS$\downarrow$          & FVD$\downarrow$             \\ \midrule
Unianimate  & 1.44E-4          & 10.05          & 0.325          & 0.617          & 1385.64         \\
ControlNeXt & 1.55E-4          & 9.84           & 0.296          & 0.620          & 1709.36         \\
Animate-X   & 1.37E-4          & 10.45          & 0.368          & 0.592          & 1267.13         \\ \midrule
Ours        & \textbf{1.05E-4} & \textbf{14.13} & \textbf{0.488} & \textbf{0.425} & \textbf{830.10} \\ \bottomrule
\end{tabular}
}
\end{center}
\label{table:cartoon}
\vspace{-0.20in}
\end{table}

\noindent \textbf{Speed.}
We compare our StableAnimator++ with current human image animation models in terms of inference latency and GPU memory consumption. Table \ref{table:inference_latency} describes the comparison results. The inference latency and GPU memory consumption are measured when the model generates 16 frames at a resolution of 576×1024. We can observe that StableAnimator++ achieves better results at a faster speed with nearly the same GPU memory consumption as AnimateAnyone~\cite{hu2024animate}, demonstrating that our model is the best trade-off between efficiency and performance. 

\begin{figure*}[t!]
\begin{center}
\includegraphics[width=0.98\linewidth]{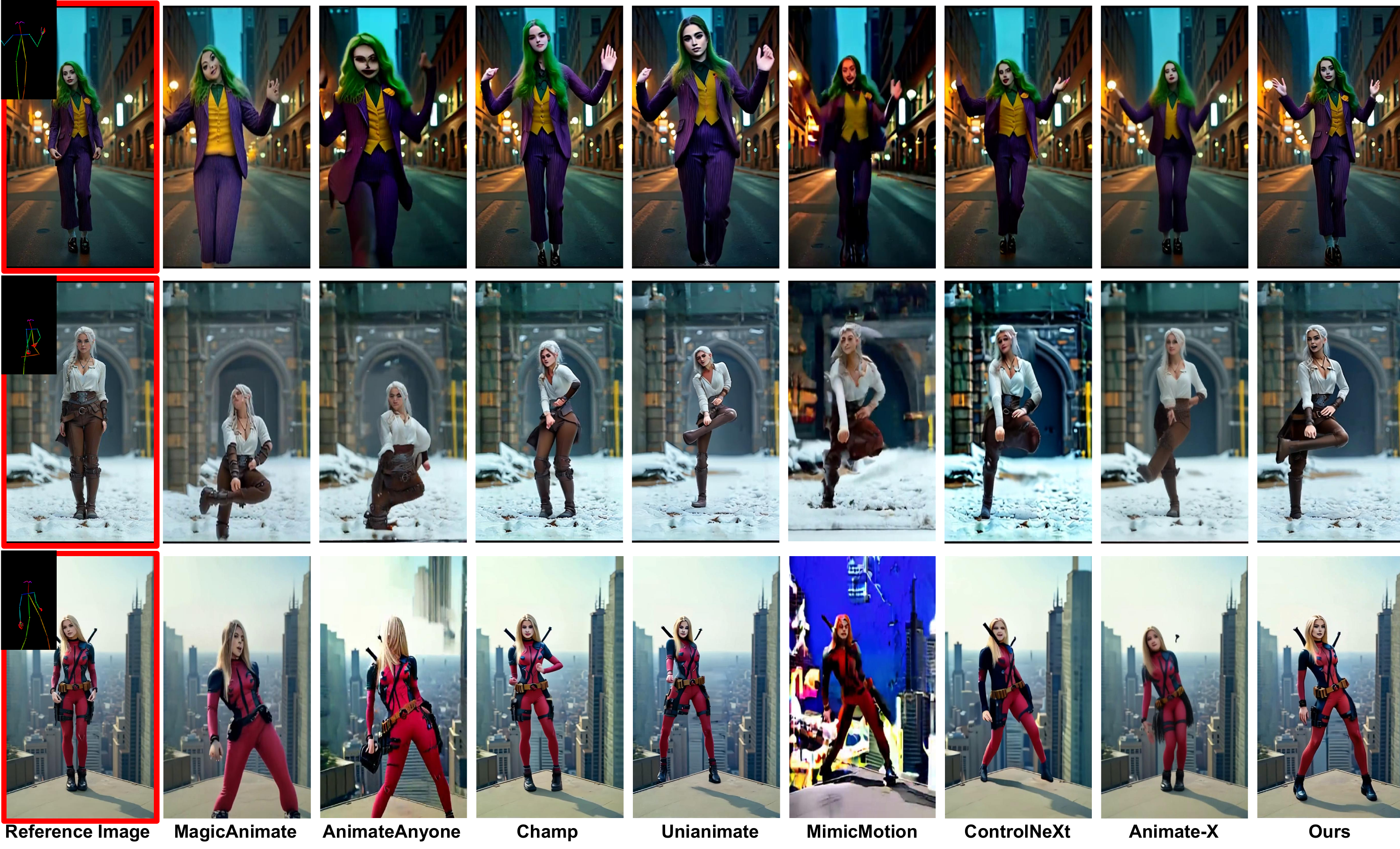}
\end{center}
\vspace{-0.55cm}
   \caption{Long animation results. The presented skeletons are misaligned with the reference image in body size and position.}
\label{fig:main_long_video}
\vspace{-0.45cm}
\end{figure*}

\begin{figure}[t!]
\begin{center}
\includegraphics[width=1\linewidth]{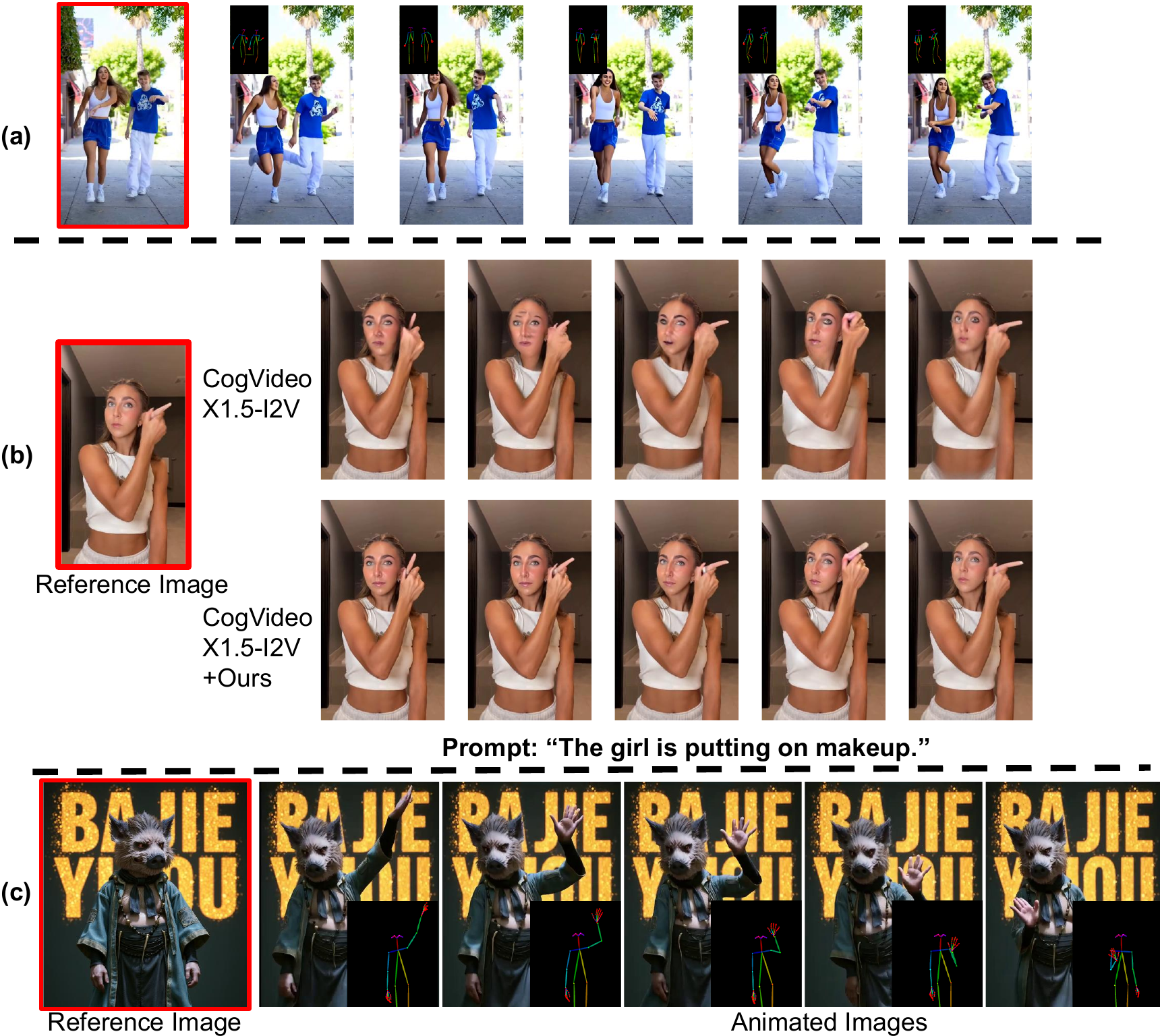}
\end{center}
\vspace{-0.6cm}
   \caption{(a), (b), and (c) refer to multiple-person animation results, general portrait generation results, and 
   anthropomorphic character animation results, respectively. The image with the red border is the reference image.
   }
\label{fig:combination}
\vspace{-0.45cm}
\end{figure}

\subsection{Application and User Study}

\noindent \textbf{Long Animation.}
We conduct qualitative comparisons between StableAnimator++ and current animation models in long animation generation, as shown in Fig. \ref{fig:main_long_video}.
Detailed comparisons are shown in Sec.VI of the Supp. Following MimicMotion~\cite{zhang2024mimicmotion}, we follow the same pipeline to synthesize long animations. Each driven pose sequence consists of over 500 frames with complex motion, and the references show significant misalignment in terms of the protagonists' body sizes and positions relative to the driven poses. We can see that competitors encounter serious body distortion and blurry noise. By contrast, our model can effectively handle long animation in high fidelity while preserving identities even in scenarios involving dramatic misalignment.

\noindent \textbf{Multi-Person Animation.}
We experiment with multiple-person animation, as shown in Fig. \ref{fig:combination} (a). The results show that our model can animate multiple people.

\noindent \textbf{General Text-to-Video Portrait Generation.}
To further validate the robustness of our core components, we integrate our face-related components (Face Encoder, ID-Adapter, and Face Optimization) into CogVideoX-I2V~\cite{yang2024cogvideox} to enable Text-to-Video generation, as shown in Fig. \ref{fig:combination} (b), indicating that our core components effectively enable the base model to maintain identity consistency without compromising video fidelity.

\noindent \textbf{Anthropomorphic Characters.}
We experiment with anthropomorphic characters, as shown in Table \ref{table:cartoon} and Fig. \ref{fig:combination}(c). As Animate-X does not release their $\bm{A}^{2}\text{Bench}$~\cite{tan2024animate_x}, we follow its method and use Kling AI to synthesize 100 anthropomorphic character videos for evaluation. We observe that ours outperforms current human image animation models.

\noindent \textbf{User Study.}
We conducted a user study with 30 video-reference image pairs to evaluate human preferences between our model and competitors. The participants are roughly university students and faculty. In each case, participants are first shown a reference image and a pose sequence with significant misalignment. Then we present two videos (one is synthesized by StableAnimator++ and the other is generated by a competitor) in random order. Participants are asked to answer the questions: M-A/A-A/B-A: ``Which one has better motion/appearance/background alignment with the reference". Table \ref{table:user_study} shows the superiority of our model in subjective evaluation.

\balance

\section{Conclusion}
We propose StableAnimator++, a robust video diffusion model with dedicated training and inference modules for generating ID-preserving human animations, even under pose misalignment. It first uses SVD-guided learnable layers to predict transformation matrices that align driven poses, significantly reducing the body size and position gap with the reference. 
StableAnimator++ then used off-the-shelf models to gain image and face embeddings.
To capture the global context of the reference, StableAnimator introduced a Face Encoder to refine face embeddings. 
An ID-Adapter then performs distribution alignment to mitigate temporal interference, enabling seamless face embedding integration without degrading video fidelity. During inference, a hybrid of the HJB equation and diffusion denoising further enhances face quality. Experiments on multiple datasets demonstrate the model’s superiority in generating high-quality, ID-consistent animations, even in misalignment scenarios.

\begin{table}[t!]\small
\caption{User preference of AnimateMaster compared with other competitors. Higher indicates users prefer more to our model.
}
\begin{center}
\renewcommand\arraystretch{1.1}
\scalebox{0.9}{
\begin{tabular}{lccc}
\toprule
Model         & M-A    & A-A    & B-A    \\ \midrule
MagicAnimate~\cite{xu2024magicanimate}  & 95.7\% & 98.5\% & 93.4\% \\
AnimateAnyone~\cite{hu2024animate} & 94.8\% & 98.2\% & 92.3\% \\
Champ~\cite{zhu2024champ}         & 92.3\% & 95.6\% & 91.8\% \\
Unianimate~\cite{wang2024unianimate}    & 91.2\% & 95.8\% & 90.6\% \\
MimicMotion~\cite{zhang2024mimicmotion}   & 90.6\% & 96.9\% & 91.5\% \\
ControlNeXt~\cite{peng2024controlnext}   & 88.6\% & 93.1\% & 90.2\% \\
Animate-X~\cite{tan2024animate_x}     & 92.4\% & 92.2\% & 90.7\% \\ \bottomrule
\end{tabular}
}
\end{center}
\label{table:user_study}
\vspace{-0.2in}
\end{table}

{
\bibliographystyle{IEEEtran}
\bibliography{egbib}
}

\begin{IEEEbiography}[{\includegraphics[width=1.2in,height=1.25in,clip,keepaspectratio]{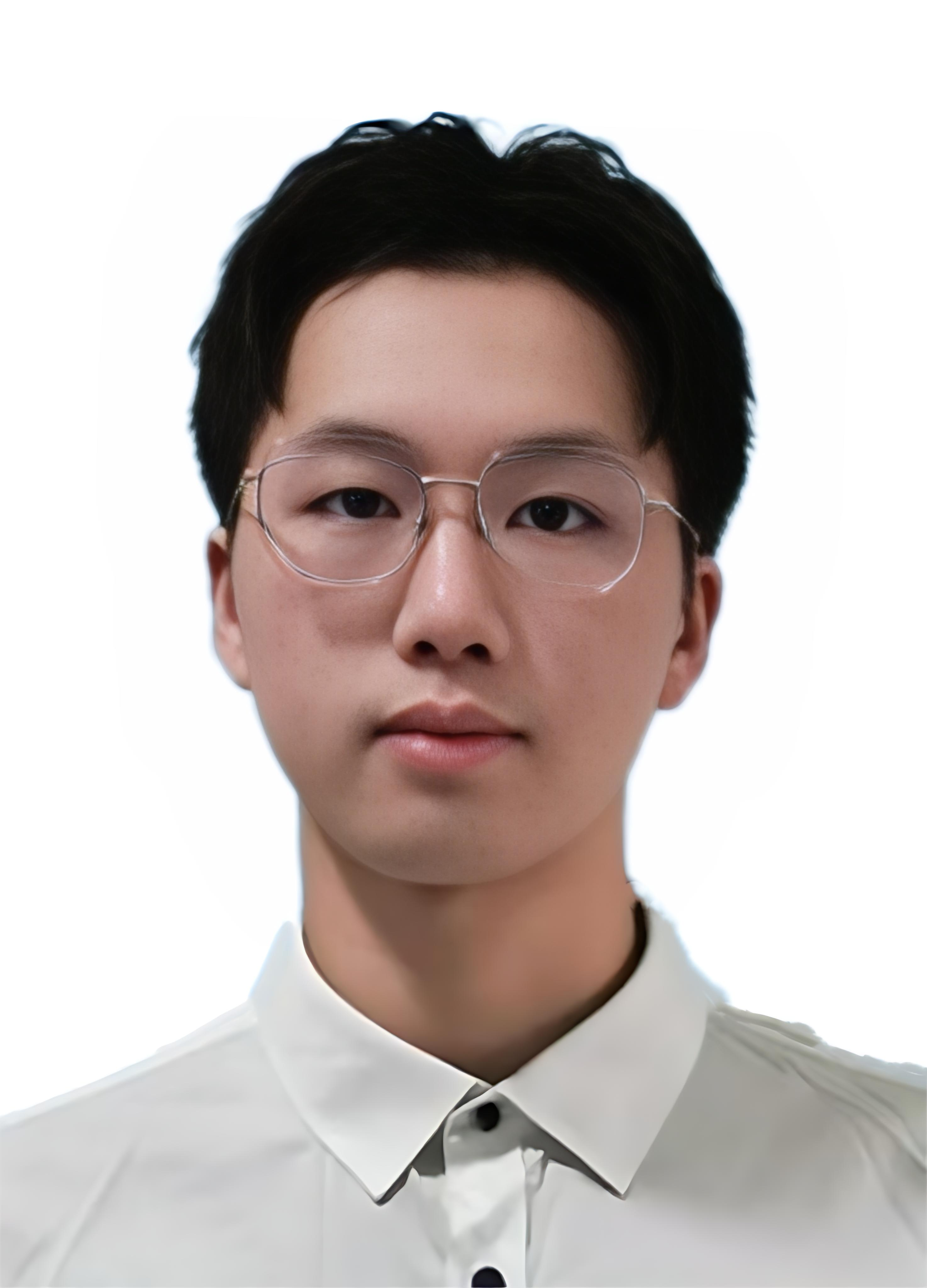}}]{Shuyuan Tu} is currently pursuing his Ph.D. degree in Computer Science in Fudan University. His research interests include conditioned video generation and video understanding. He has published several papers in top-tier conferences such as ICCV and CVPR, focusing on video generation and understanding, and is the main contributor and leader of the open-source model StableAnimator~\cite{tu2024stableanimator} (Github Star 1.3K+) for conditioned video generation.
\end{IEEEbiography}

\begin{IEEEbiography}
[{\includegraphics[width=1.2in,height=1.25in,clip,keepaspectratio]{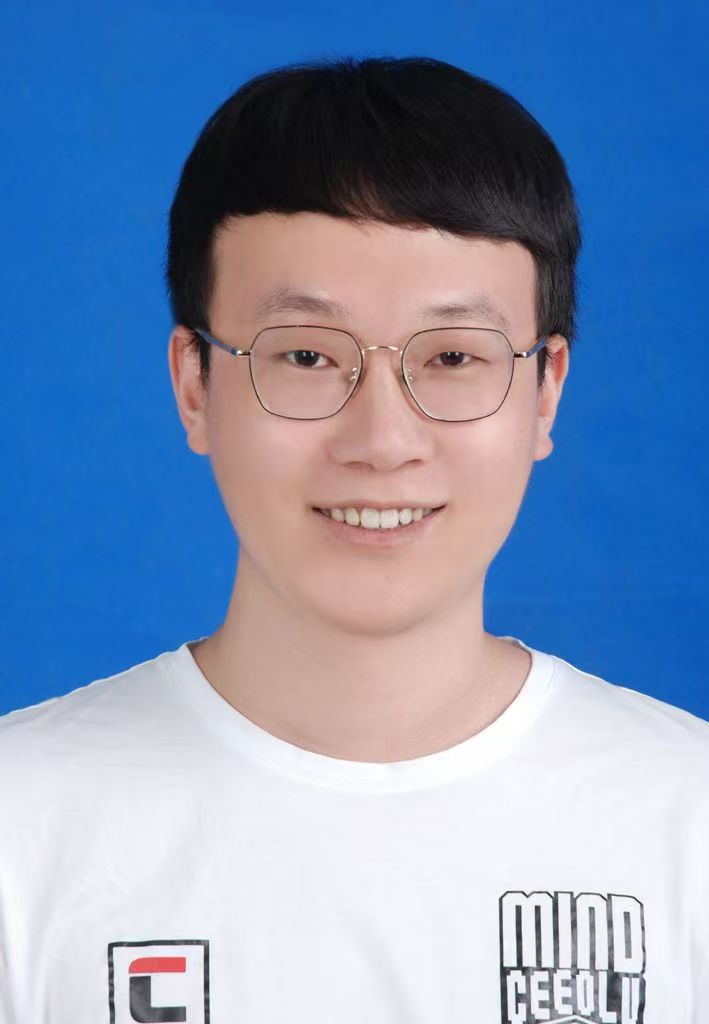}}]{Zhen Xing}
received his Ph.D. degree in Computer Science from Fudan University and has published over 20 papers in top AI conferences such as CVPR, ICCV, and ECCV. His research focuses on generative models and video understanding. He joined Alibaba Tongyi Lab in 2025 and is now a Research Scientist, focusing on video generation models.
\end{IEEEbiography}

\begin{IEEEbiography}
[{\includegraphics[width=1.2in,height=1.25in,clip,keepaspectratio]{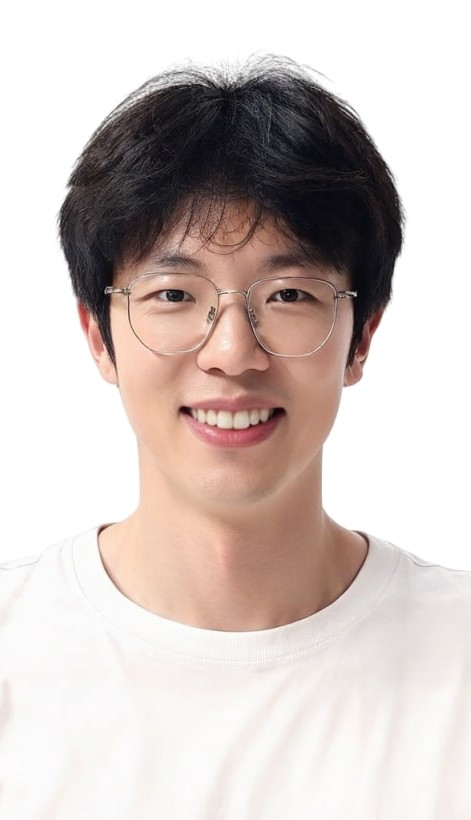}}]{Xintong Han} is currently a Senior Researcher at Tencent Hunyuan. From 2019 to 2025, he served as a Senior Tech Lead Manager at Huya Inc. He received his Ph.D. degree in Electrical and Computer Engineering from the University of Maryland, College Park, MD, USA, under the supervision of Prof. Larry S. Davis. He also obtained his B.S. degree from Shanghai Jiao Tong University, Shanghai, China. His research interests include computer vision, deep learning, and multimodal understanding and generation.
\end{IEEEbiography}

\vspace{-0.5cm}

\begin{IEEEbiography}
[{\includegraphics[width=1.2in,height=1.25in,clip,keepaspectratio]{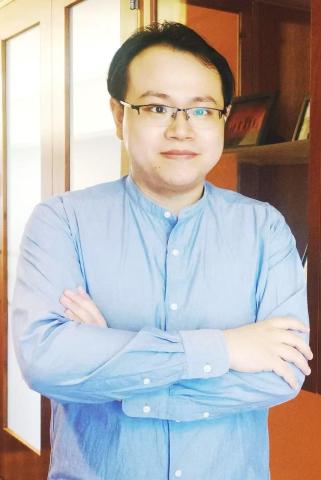}}]{Zhi-Qi Cheng} was a project scientist at Carnegie Mellon University’s Language Technologies Institute (School of Computer Science).
He is currently an assistant professor at the University of Washington. He also serve as a part‑time Research Scientist with the Meta AI AGI team.
His research interests include Multimodal Generative AI, Embodied/Robotic Intelligence, and Intelligent Transportation.
\end{IEEEbiography}

\vspace{-0.5cm}

\begin{IEEEbiography}
[{\includegraphics[width=1.2in,height=1.25in,clip,keepaspectratio]{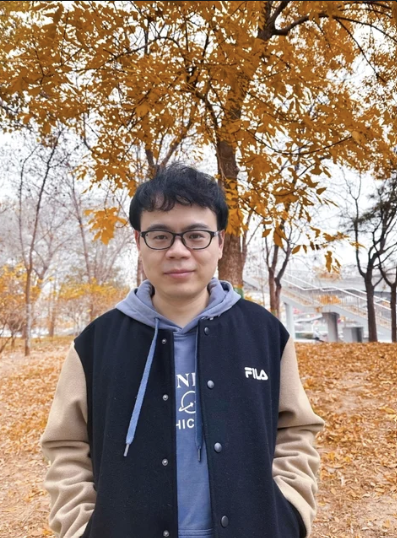}}]{Qi Dai} (Member, IEEE) received the Ph.D. degree in computer science from Fudan University, China, in 2017. He is currently a Principal Researcher with Microsoft Research, Beijing. His research interests include image/video understanding, video generation, and multimedia.
\end{IEEEbiography}

\begin{IEEEbiography}
[{\includegraphics[width=1.2in,height=1.25in,clip,keepaspectratio]{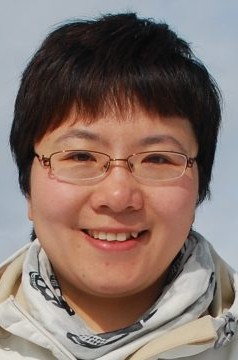}}]{Chong Luo} (Senior Member, IEEE) received her B.Sc. degree from Fudan University in Shanghai, China, in 2000, her M.S. degree from the National University of Singapore in 2002 and her Ph.D. degree from Shanghai Jiao Tong University in 2012. She joined Microsoft Research Asia (MSRA) in 2003 and is now a Senior Principal Research Manager with the Visual Computing Group. Her current research interests include image/video understanding, video generation and editing, and intelligent multimedia systems. Her work has been recognized with the ICLR 2023 Outstanding Paper Award.
\end{IEEEbiography}

\vspace{-0.5cm}

\begin{IEEEbiography}
[{\includegraphics[width=1.2in,height=1.25in,clip,keepaspectratio]{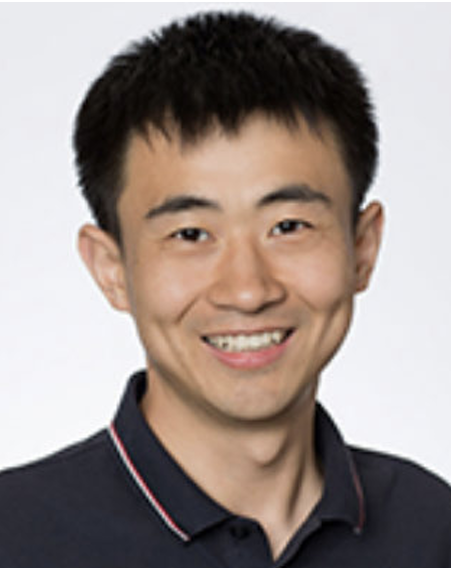}}]{Zuxuan Wu} (Member, IEEE) received the Ph.D. degree in computer science from the University of Maryland, College Park, MD, USA, with Prof. Larry S. Davis, in 2020. He is currently an Associate Professor with the Institute of Trustworthy Embodied AI, Fudan University, Shanghai, China. His research interests are in computer vision and deep learning. Dr. Wu’s work has been recognized by the AI 2000 Most Influential Scholars Honorable Mention in 2021, the Microsoft Research Ph.D. Fellowship (ten people worldwide) in 2019, and the Snap Ph.D. Fellowship (ten people worldwide) in 2017.
\end{IEEEbiography}

\vspace{-0.5cm}

\begin{IEEEbiography}
[{\includegraphics[width=1.2in,height=1.25in,clip,keepaspectratio]{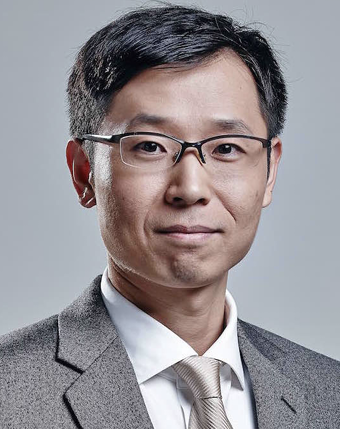}}]{Yu-Gang Jiang} (Fellow, IEEE) received the Ph.D. degree in computer science from the City University of Hong Kong, Hong Kong, in 2009. He worked as a Postdoctoral Research Scientist with Columbia University, New York, NY, USA, from 2009 to 2011. He is currently a Professor with the Institute of Trustworthy Embodied AI, Fudan University, Shanghai, China. His research lies in the areas of multimedia, compute vision, and robust and trustworthy AI. Dr. Jiang’s work has led to many awards, including the inaugural ACM China Rising Star Award, the 2015 ACM SIGMM Rising Star Award, the Research Award for Excellent Young Scholars from NSF China, and the Chang Jiang Distinguished Professorship appointed by the Ministry of Education of China.
\end{IEEEbiography}

\end{document}